\documentclass[graybox]{svmult}
\usepackage{makeidx}         % allows index generation
\usepackage{graphicx}        % standard LaTeX graphics tool 
\makeindex                   % used for the subject index 

\usepackage{amsmath,amssymb, amsfonts}        
\let\emptyset\varnothing 
 
\usepackage{xspace}
\usepackage{color}
\usepackage{epsfig}    
\graphicspath{{.}{./}}

\usepackage{tikzfig}
\usepackage{keycommand}

\newkeycommand{\boxmap}[style=small box][1]{\,\tikz{\node[style=\commandkey{style}] (x) {$#1$};\draw(0,-1.25)--(x)--(0,1.25);}\,}
\newkeycommand{\dboxmap}[style=dbox][1]{\,\tikz{\node[style=\commandkey{style}] (x) {$#1$};\draw[boldedge] (0,-1.25)--(x)--(0,1.25);}\,}

\newkeycommand{\wirelabel}[style=white label]{\,\tikz{\node[style=\commandkey{style}]{};}\,\xspace}

\newkeycommand{\pointmap}[style=point][1]{\,\tikz{\node[style=\commandkey{style}] (x) at (0,-0.3) {$#1$};\node [style=none] (3) at (0, 0.9) {};\node [style=none] (3) at (0, -0.9) {};\draw (x)--(0,0.7);}\,}

\newkeycommand{\point}[style=point,estyle=][1]{\,\tikz{\node[style=\commandkey{style}] (x) at (0,-0.05) {$#1$};\draw [\commandkey{estyle}] (x)--(0,1.05);}\,}

\newkeycommand{\copointmap}[style=copoint][1]{\,\tikz{\node[style=\commandkey{style}] (x) at (0,0.3) {$#1$};\draw (x)--(0,-0.7);}\,}

\newkeycommand{\dpoint}[style=point][1]{\,\tikz{\node[style=\commandkey{style},doubled] (x) at (0,-0.3) {$#1$};\draw[boldedge] (x)--(0,0.7);}\,}

\newkeycommand{\kpoint}[style=kpoint,estyle=][1]{\,\tikz{\node[style=\commandkey{style}] (x) at (0,-0.05) {$#1$};\draw [\commandkey{estyle}] (x)--(0,1.05);}\,}

\newkeycommand{\kpointgray}[style=gray kpoint,estyle=][1]{\,\tikz{\node[style=\commandkey{style}] (x) at (0,-0.05) {$#1$};\draw [\commandkey{estyle}] (x)--(0,1.05);}\,}

\newkeycommand{\typedkpoint}[style=kpoint,edgestyle=][2]{%
\,\begin{tikzpicture}
  \begin{pgfonlayer}{nodelayer}
    \node [style=none] (0) at (0, 1) {};
    \node [style=\commandkey{style}] (1) at (0, -0.75) {$#2$};
    \node [style=right label] (2) at (0.25, 0.5) {$#1$};
  \end{pgfonlayer}
  \begin{pgfonlayer}{edgelayer}
    \draw [\commandkey{edgestyle}] (1) to (0.center);
  \end{pgfonlayer}
\end{tikzpicture}\,}

\newkeycommand{\typedkpointdag}[style=kpoint adjoint,edgestyle=][2]{%
\,\begin{tikzpicture}
  \begin{pgfonlayer}{nodelayer}
    \node [style=none] (0) at (0, -1) {};
    \node [style=\commandkey{style}] (1) at (0, 0.75) {$#2$};
    \node [style=right label] (2) at (0.25, -0.5) {$#1$};
  \end{pgfonlayer}
  \begin{pgfonlayer}{edgelayer}
    \draw [\commandkey{edgestyle}] (0.center) to (1);
  \end{pgfonlayer}
\end{tikzpicture}\,}

\newkeycommand{\bistate}[style=kpoint][1]{\,\begin{tikzpicture}
  \begin{pgfonlayer}{nodelayer}
    \node [style=\commandkey{style}, minimum width=1 cm, inner sep=2pt] (0) at (0, -0.25) {$#1$};
    \node [style=none] (1) at (-0.75, 0) {};
    \node [style=none] (2) at (0.75, 0) {};
    \node [style=none] (3) at (-0.75, 0.88) {};
    \node [style=none] (4) at (0.75, 0.88) {};
  \end{pgfonlayer}
  \begin{pgfonlayer}{edgelayer}
    \draw (1.center) to (3.center);
    \draw (2.center) to (4.center);
  \end{pgfonlayer}
\end{tikzpicture}\,}

\newkeycommand{\bistatebig}[style=kpoint][1]{\,\begin{tikzpicture}
  \begin{pgfonlayer}{nodelayer}
    \node [style=\commandkey{style}, minimum width=, minimum width=1.26 cm, inner sep=2pt] (0) at (0, -0.25) {$#1$};
    \node [style=none] (1) at (-1, 0) {};
    \node [style=none] (2) at (1, 0) {};
    \node [style=none] (3) at (-1, 0.88) {};
    \node [style=none] (4) at (1, 0.88) {};
  \end{pgfonlayer}
  \begin{pgfonlayer}{edgelayer}
    \draw (1.center) to (3.center);
    \draw (2.center) to (4.center);
  \end{pgfonlayer}
\end{tikzpicture}\,}

\newkeycommand{\dbistate}[style=dkpoint][1]{\,\begin{tikzpicture}
  \begin{pgfonlayer}{nodelayer}
    \node [style=\commandkey{style}, minimum width=1 cm, inner sep=2pt] (0) at (0, -0.25) {$#1$};
    \node [style=none] (1) at (-0.75, 0) {};
    \node [style=none] (2) at (0.75, 0) {};
    \node [style=none] (3) at (-0.75, 1.0) {};
    \node [style=none] (4) at (0.75, 1.0) {};
  \end{pgfonlayer}
  \begin{pgfonlayer}{edgelayer}
    \draw [boldedge] (1.center) to (3.center);
    \draw [boldedge] (2.center) to (4.center);
  \end{pgfonlayer}
\end{tikzpicture}\,}

\newkeycommand{\bistatebraket}[style1=kpoint,style2=kpointadj][2]{\,\begin{tikzpicture}
  \begin{pgfonlayer}{nodelayer}
    \node [style=\commandkey{style1}, minimum width=1 cm, inner sep=2pt] (0) at (0, -0.75) {$#1$};
    \node [style=none] (1) at (-0.75, -0.5) {};
    \node [style=none] (2) at (0.75, -0.5) {};
    \node [style=none] (3) at (-0.75, 0.5) {};
    \node [style=none] (4) at (0.75, 0.5) {};
    \node [style=\commandkey{style2}, minimum width=1 cm, inner sep=2pt] (5) at (0, 0.75) {$#2$};
  \end{pgfonlayer}
  \begin{pgfonlayer}{edgelayer}
    \draw (1.center) to (3.center);
    \draw (2.center) to (4.center);
  \end{pgfonlayer}
\end{tikzpicture}\,}

\newkeycommand{\weight}[style=dkpoint][1]{\,\begin{tikzpicture}
  \begin{pgfonlayer}{nodelayer}
    \node [style=\commandkey{style}] (0) at (0, -0.5) {$#1$};
    \node [style=upground] (1) at (0, 0.75) {};
  \end{pgfonlayer}
  \begin{pgfonlayer}{edgelayer}
    \draw [style=boldedge] (0) to (1);
  \end{pgfonlayer}
\end{tikzpicture}\,}

\newkeycommand{\dkpoint}[style=dkpoint][1]{\,\tikz{\node[style=\commandkey{style}] (x) at (0,-0.1) {$#1$};\draw[boldedge] (x)--(0,1);}\,}
\newkeycommand{\dkpointadj}[style=dkpointadj][1]{\,\tikz{\node[style=\commandkey{style}] (x) at (0,0.1) {$#1$};\draw[boldedge] (x)--(0,-1);}\,}
\newkeycommand{\dkpointtrans}[style=dkpointtrans][1]{\,\tikz{\node[style=\commandkey{style}] (x) at (0,0.1) {$#1$};\draw[boldedge] (x)--(0,-1);}\,}
\newkeycommand{\dkpointconj}[style=dkpointconj][1]{\,\tikz{\node[style=\commandkey{style}] (x) at (0,-0.1) {$#1$};\draw[boldedge] (x)--(0,1);}\,}

\newkeycommand{\blackkpoint}[style=black kpoint][1]{\,\tikz{\node[style=\commandkey{style}] (x) at (0,-0.1) {$#1$};\draw (x)--(0,1);}\,}
\newkeycommand{\blackkpointadj}[style=black kpointadj][1]{\,\tikz{\node[style=\commandkey{style}] (x) at (0,0.1) {$#1$};\draw (x)--(0,-1);}\,}
\newkeycommand{\blackdkpoint}[style=black dkpoint][1]{\,\tikz{\node[style=\commandkey{style}] (x) at (0,-0.1) {$#1$};\draw[boldedge] (x)--(0,1);}\,}
\newkeycommand{\blackdkpointadj}[style=black dkpointadj][1]{\,\tikz{\node[style=\commandkey{style}] (x) at (0,0.1) {$#1$};\draw[boldedge] (x)--(0,-1);}\,}

\newcommand{\trace}{\,\begin{tikzpicture}
  \begin{pgfonlayer}{nodelayer}
    \node [style=none] (0) at (0, -0.60) {};
    \node [style=upground] (1) at (0, 0.40) {};
  \end{pgfonlayer}
  \begin{pgfonlayer}{edgelayer}
    \draw [style=none] (0.center) to (1);
  \end{pgfonlayer}
\end{tikzpicture}\,}

\newcommand\discard\trace

\newkeycommand{\pointketbra}[style1=copoint,style2=point,estyle=][2]{\,%
\begin{tikzpicture}
  \begin{pgfonlayer}{nodelayer}
    \node [style=none] (0) at (0, -1.75) {};
    \node [style=\commandkey{style1}] (1) at (0, -0.75) {$#1$};
    \node [style=\commandkey{style2}] (2) at (0, 0.75) {$#2$};
    \node [style=none] (3) at (0, 1.75) {};
  \end{pgfonlayer}
  \begin{pgfonlayer}{edgelayer}
    \draw [\commandkey{estyle}] (0.center) to (1);
    \draw [\commandkey{estyle}] (2) to (3.center);
  \end{pgfonlayer}
\end{tikzpicture}\,}

\newkeycommand{\twocopointketbra}[style1=copoint,style2=copoint,style3=point][3]{\,%
\begin{tikzpicture}
  \begin{pgfonlayer}{nodelayer}
    \node [style=none] (0) at (-0.75, -1.75) {};
    \node [style=none] (1) at (0.75, -1.75) {};
    \node [style=\commandkey{style1}] (2) at (-0.75, -0.75) {$#1$};
    \node [style=\commandkey{style2}] (3) at (0.75, -0.75) {$#2$};
    \node [style=\commandkey{style3}] (4) at (0, 0.75) {$#3$};
    \node [style=none] (5) at (0, 1.75) {};
  \end{pgfonlayer}
  \begin{pgfonlayer}{edgelayer}
    \draw (0.center) to (2);
    \draw (1.center) to (3);
    \draw (4) to (5.center);
  \end{pgfonlayer}
\end{tikzpicture}\,}

\newkeycommand{\twopointketbra}[style1=copoint,style2=point,style3=point][3]{\,%
\begin{tikzpicture}
  \begin{pgfonlayer}{nodelayer}
    \node [style=none] (0) at (0, -1.75) {};
    \node [style=none] (1) at (-0.75, 1.75) {};
    \node [style=\commandkey{style1}] (2) at (0, -0.75) {$#1$};
    \node [style=\commandkey{style2}] (3) at (-0.75, 0.75) {$#2$};
    \node [style=\commandkey{style3}] (4) at (0.75, 0.75) {$#3$};
    \node [style=none] (5) at (0.75, 1.75) {};
  \end{pgfonlayer}
  \begin{pgfonlayer}{edgelayer}
    \draw (0.center) to (2);
    \draw (1.center) to (3);
    \draw (4) to (5.center);
  \end{pgfonlayer}
\end{tikzpicture}\,}

\newkeycommand{\pointbraket}[style1=point,style2=copoint,estyle=][2]{\,%
\begin{tikzpicture}
  \begin{pgfonlayer}{nodelayer}
    \node [style=\commandkey{style1}] (0) at (0, -0.625) {$#1$};
    \node [style=\commandkey{style2}] (1) at (0, 0.625) {$#2$};
  \end{pgfonlayer}
  \begin{pgfonlayer}{edgelayer}
    \draw [\commandkey{estyle}] (0) to (1);
  \end{pgfonlayer}
\end{tikzpicture}\,}

\newkeycommand{\kpointketbra}[style1=kpointdag,style2=kpoint,estyle=][2]{\,%
\begin{tikzpicture}
  \begin{pgfonlayer}{nodelayer}
    \node [style=none] (0) at (0, -1.9) {};
    \node [style=\commandkey{style1}] (1) at (0, -0.95) {$#1$};
    \node [style=\commandkey{style2}] (2) at (0, 0.95) {$#2$};
    \node [style=none] (3) at (0, 1.9) {};
  \end{pgfonlayer}
  \begin{pgfonlayer}{edgelayer}
    \draw [\commandkey{estyle}] (0.center) to (1);
    \draw [\commandkey{estyle}] (2) to (3.center);
  \end{pgfonlayer}
\end{tikzpicture}\,}

\newkeycommand{\kpointmap}[style1=kpoint,style2=map][2]{\,%
\begin{tikzpicture}
  \begin{pgfonlayer}{nodelayer}
    \node [style=\commandkey{style1}] (0) at (0, -1.1) {$#1$};
    \node [style=\commandkey{style2}] (1) at (0, 0.5) {$#2$};
    \node [style=none] (2) at (0, 1.75) {};
  \end{pgfonlayer}
  \begin{pgfonlayer}{edgelayer}
    \draw (0) to (1);
    \draw (1) to (2.center);
  \end{pgfonlayer}
\end{tikzpicture}%
\,}

\newkeycommand{\kpointmaptrans}[style1=kpointconj,style2=mapadj][2]{\,%
\begin{tikzpicture}
  \begin{pgfonlayer}{nodelayer}
    \node [style=\commandkey{style1}] (0) at (0, -1.1) {$#1$};
    \node [style=\commandkey{style2}] (1) at (0, 0.5) {$#2$};
    \node [style=none] (2) at (0, 1.75) {};
  \end{pgfonlayer}
  \begin{pgfonlayer}{edgelayer}
    \draw (0) to (1);
    \draw (1) to (2.center);
  \end{pgfonlayer}
\end{tikzpicture}%
\,}

\newkeycommand{\kpointmapadj}[style1=kpointadj,style2=map][2]{\,%
\begin{tikzpicture}
  \begin{pgfonlayer}{nodelayer}
    \node [style=\commandkey{style1}] (0) at (0, 1.1) {$#1$};
    \node [style=\commandkey{style2}] (1) at (0, -0.5) {$#2$};
    \node [style=none] (2) at (0, -1.75) {};
  \end{pgfonlayer}
  \begin{pgfonlayer}{edgelayer}
    \draw (0) to (1);
    \draw (1) to (2.center);
  \end{pgfonlayer}
\end{tikzpicture}%
\,}

\newkeycommand{\dkpointmap}[style1=dkpoint,style2=dmap][2]{\,%
\begin{tikzpicture}
  \begin{pgfonlayer}{nodelayer}
    \node [style=\commandkey{style1}] (0) at (0, -1.2) {$#1$};
    \node [style=\commandkey{style2}] (1) at (0, 0.4) {$#2$};
    \node [style=none] (2) at (0, 1.7) {};
  \end{pgfonlayer}
  \begin{pgfonlayer}{edgelayer}
    \draw[style=boldedge] (0) to (1);
    \draw[style=boldedge] (1) to (2.center);
  \end{pgfonlayer}
\end{tikzpicture}%
\,}

\newkeycommand{\dkpointmapx}[style1=dkpoint,style2=dmap][2]{\,%
\begin{tikzpicture}
  \begin{pgfonlayer}{nodelayer}
    \node [style=\commandkey{style1}] (0) at (0, -1.4) {$#1$};
    \node [style=\commandkey{style2}] (1) at (0, 0.6) {$#2$};
    \node [style=none] (2) at (0, 1.9) {};
  \end{pgfonlayer}
  \begin{pgfonlayer}{edgelayer}
    \draw[style=boldedge] (0) to (1);
    \draw[style=boldedge] (1) to (2.center);
  \end{pgfonlayer}
\end{tikzpicture}%
\,}

\newkeycommand{\boxcopointmapdag}[style1=map,style2=copoint][2]{\,%
\begin{tikzpicture}
  \begin{pgfonlayer}{nodelayer}
    \node [style=none] (0) at (0, -1.75) {};
    \node [style=\commandkey{style1}] (1) at (0, -0.5) {$#1$};
    \node [style=\commandkey{style2}] (2) at (0, 1) {$#2$};
  \end{pgfonlayer}
  \begin{pgfonlayer}{edgelayer}
    \draw (0.center) to (1);
    \draw (1) to (2);
  \end{pgfonlayer}
\end{tikzpicture}%
\,}

\newkeycommand{\kpointdagmap}[style1=map,style2=kpointdag][2]{\,%
\begin{tikzpicture}
  \begin{pgfonlayer}{nodelayer}
    \node [style=none] (0) at (0, -1.75) {};
    \node [style=\commandkey{style1}] (1) at (0, -0.5) {$#1$};
    \node [style=\commandkey{style2}] (2) at (0, 1.1) {$#2$};
  \end{pgfonlayer}
  \begin{pgfonlayer}{edgelayer}
    \draw (0.center) to (1);
    \draw (1) to (2);
  \end{pgfonlayer}
\end{tikzpicture}%
\,}

\newkeycommand{\sandwichmap}[style1=point,style2=map,style3=copoint][3]{\,%
\begin{tikzpicture}
  \begin{pgfonlayer}{nodelayer}
    \node [style=\commandkey{style1}] (0) at (0, -1.50) {$#1$};
    \node [style=\commandkey{style2}] (1) at (0, 0) {$#2$};
    \node [style=\commandkey{style3}] (2) at (0, 1.50) {$#3$};
  \end{pgfonlayer}
  \begin{pgfonlayer}{edgelayer}
    \draw (0) to (1);
    \draw (1) to (2);
  \end{pgfonlayer}
\end{tikzpicture}%
}

\newkeycommand{\kpointsandwichmap}[style1=kpoint,style2=map,style3=kpointdag][3]{\,%
\begin{tikzpicture}
  \begin{pgfonlayer}{nodelayer}
    \node [style=\commandkey{style1}] (0) at (0, -1.5) {$#1$};
    \node [style=\commandkey{style2}] (1) at (0, 0) {$#2$};
    \node [style=\commandkey{style3}] (2) at (0, 1.5) {$#3$};
  \end{pgfonlayer}
  \begin{pgfonlayer}{edgelayer}
    \draw (0) to (1);
    \draw (1) to (2);
  \end{pgfonlayer}
\end{tikzpicture}%
}

\newkeycommand{\sandwichmapconj}[style1=point,style2=mapconj,style3=copoint][3]{\,% 
\begin{tikzpicture}
  \begin{pgfonlayer}{nodelayer}
    \node [style=\commandkey{style1}] (0) at (0, -1.50) {$#1$};
    \node [style=\commandkey{style2}] (1) at (0, 0) {$#2$};
    \node [style=\commandkey{style3}] (2) at (0, 1.50) {$#3$};
  \end{pgfonlayer}
  \begin{pgfonlayer}{edgelayer}
    \draw (0) to (1);
    \draw (1) to (2);
  \end{pgfonlayer}
\end{tikzpicture}%
}

\newkeycommand{\sandwichtwo}[style1=point,style2=map,style3=map,style4=copoint][4]{\,%
\begin{tikzpicture}
  \begin{pgfonlayer}{nodelayer}
    \node [style=\commandkey{style2}] (0) at (0, -1) {$#2$};
    \node [style=\commandkey{style3}] (1) at (0, 1) {$#3$};
    \node [style=\commandkey{style1}] (2) at (0, -2.6) {$#1$};
    \node [style=\commandkey{style4}] (3) at (0, 2.6) {$#4$};
  \end{pgfonlayer}
  \begin{pgfonlayer}{edgelayer}
    \draw (2) to (0);
    \draw (0) to (1);
    \draw (1) to (3);
  \end{pgfonlayer}
\end{tikzpicture}\,}

\newkeycommand{\longbraket}[style1=point,style2=copoint][2]{\,% 
\begin{tikzpicture}
  \begin{pgfonlayer}{nodelayer}
    \node [style=\commandkey{style1}] (0) at (0, -2.6) {$#1$};
    \node [style=\commandkey{style2}] (1) at (0, 2.6) {$#2$};
  \end{pgfonlayer}
  \begin{pgfonlayer}{edgelayer}
    \draw (0) to (1);
  \end{pgfonlayer}
\end{tikzpicture}\,}

\newkeycommand{\onb}[style=point]{\ensuremath{\{\,\tikz{\node[style=\commandkey{style}] (x) at (0,-0.3) {$j$};\draw (x)--(0,0.7);}\,\}}\xspace}

\newkeycommand{\phase}[style=white phase dot][1]{\,\begin{tikzpicture}
    \begin{pgfonlayer}{nodelayer}
        \node [style=none] (0) at (0, 1) {};
        \node [style=\commandkey{style}] (2) at (0, -0) {$#1$}; 
        \node [style=none] (3) at (0, -1) {};
    \end{pgfonlayer}
    \begin{pgfonlayer}{edgelayer}
        \draw (2) to (0.center);
        \draw (3.center) to (2);
    \end{pgfonlayer}
\end{tikzpicture}\,}

\newkeycommand{\dphase}[style=white phase ddot][1]{\,\begin{tikzpicture}
    \begin{pgfonlayer}{nodelayer}
        \node [style=none] (0) at (0, 1.25) {};
        \node [style=\commandkey{style}] (2) at (0, -0) {$#1$};
        \node [style=none] (3) at (0, -1.25) {};
    \end{pgfonlayer}
    \begin{pgfonlayer}{edgelayer}
        \draw [boldedge] (2) to (0.center);
        \draw [boldedge] (3.center) to (2);
    \end{pgfonlayer}
\end{tikzpicture}\,}

\newkeycommand{\dphasepoint}[style=white phase ddot][1]{\,\begin{tikzpicture}[yshift=-3mm]
  \begin{pgfonlayer}{nodelayer}
    \node [style=none] (0) at (0, 1) {};
    \node [style=\commandkey{style}] (1) at (0, 0) {$#1$};
  \end{pgfonlayer}
  \begin{pgfonlayer}{edgelayer}
    \draw [style=boldedge] (0.center) to (1);
  \end{pgfonlayer}
\end{tikzpicture}\,}

\newkeycommand{\dphasepointgray}[style=gray phase ddot][1]{\,\begin{tikzpicture}[yshift=-3mm]
  \begin{pgfonlayer}{nodelayer}
    \node [style=none] (0) at (0, 1) {};
    \node [style=\commandkey{style}] (1) at (0, 0) {$#1$};
  \end{pgfonlayer}
  \begin{pgfonlayer}{edgelayer}
    \draw [style=boldedge] (0.center) to (1);
  \end{pgfonlayer}
\end{tikzpicture}\,}

\newkeycommand{\dphasecopoint}[style=white phase ddot][1]{\,\begin{tikzpicture}[yshift=3mm,yscale=-1]
  \begin{pgfonlayer}{nodelayer}
    \node [style=none] (0) at (0, 1) {};
    \node [style=\commandkey{style}] (1) at (0, 0) {$#1$};
  \end{pgfonlayer}
  \begin{pgfonlayer}{edgelayer}
    \draw [style=boldedge] (0.center) to (1);
  \end{pgfonlayer}
\end{tikzpicture}\,}

\newkeycommand{\dphasepointsm}[style=white phase ddot][1]{\,\begin{tikzpicture}[yshift=-3mm]
  \begin{pgfonlayer}{nodelayer}
    \node [style=none] (0) at (0, 1) {};
    \node [style=\commandkey{style}] (1) at (0, 0) {\footnotesize\!$#1$\!};
  \end{pgfonlayer}
  \begin{pgfonlayer}{edgelayer}
    \draw [style=boldedge] (0.center) to (1);
  \end{pgfonlayer}
\end{tikzpicture}\,}

\newkeycommand{\phasepoint}[style=white phase dot][1]{\,\begin{tikzpicture}[yshift=-3mm]
  \begin{pgfonlayer}{nodelayer}
    \node [style=none] (0) at (0, 1) {};
    \node [style=\commandkey{style}] (1) at (0, 0) {$#1$};
  \end{pgfonlayer}
  \begin{pgfonlayer}{edgelayer}
    \draw (0.center) to (1);
  \end{pgfonlayer}
\end{tikzpicture}\,}

\newkeycommand{\phasecopoint}[style=white phase dot][1]{\,\begin{tikzpicture}[yshift=3mm]
  \begin{pgfonlayer}{nodelayer}
    \node [style=none] (0) at (0, -1) {};
    \node [style=\commandkey{style}] (1) at (0, 0) {$#1$};
  \end{pgfonlayer}
  \begin{pgfonlayer}{edgelayer}
    \draw (0.center) to (1);
  \end{pgfonlayer}
\end{tikzpicture}\,}

\newkeycommand{\kpointbraket}[style1=kpoint,style2=kpoint adjoint,estyle=][2]{\,%
\begin{tikzpicture}
  \begin{pgfonlayer}{nodelayer}
    \node [style=\commandkey{style1}] (0) at (0, -0.735) {$#1$};
    \node [style=\commandkey{style2}] (1) at (0, 0.735) {$#2$};
  \end{pgfonlayer}
  \begin{pgfonlayer}{edgelayer}
    \draw [\commandkey{estyle}] (0) to (1);
  \end{pgfonlayer}
\end{tikzpicture}\,}

\newkeycommand{\kkpointbraket}[style1=kpoint,style2=kpoint adjoint,estyle=][2]{\,%
\begin{tikzpicture}
  \begin{pgfonlayer}{nodelayer}
    \node [style=\commandkey{style1}] (0) at (0, -0.685) {$#1$};
    \node [style=\commandkey{style2}] (1) at (0, 0.685) {$#2$};
  \end{pgfonlayer}
  \begin{pgfonlayer}{edgelayer}
    \draw [\commandkey{estyle}] (0) to (1);
  \end{pgfonlayer}
\end{tikzpicture}\,}

\newkeycommand{\kpointbraketx}[style1=kpoint,style2=kpoint adjoint,estyle=][2]{\,%
\begin{tikzpicture}
  \begin{pgfonlayer}{nodelayer}
    \node [style=\commandkey{style1}] (0) at (0, -0.885) {$#1$};
    \node [style=\commandkey{style2}] (1) at (0, 0.885) {$#2$};
  \end{pgfonlayer}
  \begin{pgfonlayer}{edgelayer}
    \draw [\commandkey{estyle}] (0) to (1);
  \end{pgfonlayer}
\end{tikzpicture}\,}

\tikzstyle{cdiag}=[matrix of math nodes, row sep=3em, column sep=3em, text height=1.5ex, text depth=0.25ex,inner sep=0.5em]
\tikzstyle{arrow above}=[transform canvas={yshift=0.5ex}]
\tikzstyle{arrow below}=[transform canvas={yshift=-0.5ex}]

\usepackage{tikz}
\usetikzlibrary{decorations.markings}
\usetikzlibrary{shapes.geometric}
\pgfdeclarelayer{edgelayer}
\pgfdeclarelayer{nodelayer}
\pgfsetlayers{edgelayer,nodelayer,main}

\tikzstyle{none}=[inner sep=0pt]
\definecolor{hexcolor0xff0000}{rgb}{1.000,0.000,0.000}
\definecolor{hexcolor0x000000}{rgb}{0.000,0.000,0.000}
\definecolor{hexcolor0x00ff00}{rgb}{0.000,1.000,0.000}
\definecolor{hexcolor0x000000}{rgb}{0.000,0.000,0.000}
\definecolor{hexcolor0xffff00}{rgb}{1.000,1.000,0.000}
\definecolor{hexcolor0xffffff}{rgb}{1.000,1.000,1.000}

\tikzstyle{rn}=[circle,fill=hexcolor0xff0000,draw=hexcolor0x000000,line width=0.8 pt]
\tikzstyle{gn}=[circle,fill=hexcolor0x00ff00,draw=hexcolor0x000000,line width=0.8 pt]
\tikzstyle{yn}=[circle,fill=hexcolor0xffff00,draw=hexcolor0x000000,line width=0.8 pt]
\tikzstyle{wn}=[circle,fill=hexcolor0xffffff,draw=hexcolor0x000000,line width=0.8 pt]
\tikzstyle{wnthick}=[circle,fill=hexcolor0xffffff,draw=hexcolor0x000000,line width=2.500]

\tikzstyle{simple}=[-,draw=hexcolor0x000000,line width=2.000]
\tikzstyle{arrow}=[-,draw=hexcolor0x000000,postaction={decorate},decoration={markings,mark=at position .5 with {\arrow{>}}},line width=2.000]
\tikzstyle{tick}=[-,draw=hexcolor0x000000,postaction={decorate},decoration={markings,mark=at position .5 with {\draw (0,-0.1) -- (0,0.1);}},line width=2.000]
\tikzstyle{halfthickness}=[-,draw=hexcolor0x000000,line width=0.500]
\tikzstyle{thick}=[-,draw=hexcolor0x000000,line width=2.500]
\tikzstyle{thicker}=[-,draw=hexcolor0x000000,line width=4.000]

\tikzstyle{env}=[copoint,regular polygon rotate=0,minimum width=0.2cm, fill=black]

\tikzstyle{probs}=[shape=semicircle,fill=white,draw=black,shape border rotate=180,minimum width=1.2cm]

\tikzstyle{every picture}=[baseline=-0.25em,scale=0.5]
\tikzstyle{dotpic}=[] % for backwards-compatibility
\tikzstyle{diredges}=[every to/.style={diredge}]
\tikzstyle{math matrix}=[matrix of math nodes,left delimiter=(,right delimiter=),inner sep=2pt,column sep=1em,row sep=0.5em,nodes={inner sep=0pt},text height=1.5ex, text depth=0.25ex]

\tikzstyle{inline text}=[text height=1.5ex, text depth=0.25ex,yshift=0.5mm]
\tikzstyle{label}=[font=\footnotesize,text height=1.5ex, text depth=0.25ex,yshift=0.5mm]
\tikzstyle{left label}=[label,anchor=east,xshift=1.5mm]
\tikzstyle{right label}=[label,anchor=west,xshift=-1.5mm]

\tikzstyle{braceedge}=[decorate,decoration={brace,amplitude=2mm,raise=-1mm}]
\tikzstyle{small braceedge}=[decorate,decoration={brace,amplitude=1mm,raise=-1mm}]

\tikzstyle{doubled}=[line width=1.6pt] % set the line width for all doubled (quantum) maps/wires
\tikzstyle{boldedge}=[doubled,shorten <=-0.17mm,shorten >=-0.17mm]
\tikzstyle{boldedgegray}=[doubled,gray,shorten <=-0.17mm,shorten >=-0.17mm]

\tikzstyle{semidoubled}=[line width=1.4pt] % set the line width for all doubled (quantum) maps/wires
\tikzstyle{semiboldedgegray}=[semidoubled,gray,shorten <=-0.17mm,shorten >=-0.17mm]

\tikzstyle{boldedgedashed}=[very thick,dashed,shorten <=-0.17mm,shorten >=-0.17mm]
\tikzstyle{vboldedgedashed}=[doubled,dashed,shorten <=-0.17mm,shorten >=-0.17mm]
\tikzstyle{left hook arrow}=[left hook-latex]
\tikzstyle{right hook arrow}=[right hook-latex]
\tikzstyle{sembracket}=[line width=0.5pt,shorten <=-0.07mm,shorten >=-0.07mm]

\tikzstyle{causal edge}=[->,thick,gray]
\tikzstyle{causal nondir}=[thick,gray]
\tikzstyle{timeline}=[thick,gray, dashed]

\tikzstyle{cedge}=[<->,thick,gray!70!white]

\tikzstyle{empty diagram}=[draw=gray!40!white,dashed,shape=rectangle,minimum width=1cm,minimum height=1cm]
\tikzstyle{empty diagram small}=[draw=gray!50!white,dashed,shape=rectangle,minimum width=0.6cm,minimum height=0.5cm]

\tikzstyle{dot}=[inner sep=0mm,minimum width=2mm,minimum height=2mm,draw,shape=circle]
\tikzstyle{ddot}=[inner sep=0mm, doubled, minimum width=2.5mm,minimum height=2.5mm,draw,shape=circle]

\tikzstyle{black dot}=[dot,fill=black]
\tikzstyle{white dot}=[dot,fill=white,,text depth=-0.2mm]
\tikzstyle{green dot}=[white dot] % for backwards-compatibility
\tikzstyle{gray dot}=[dot,fill=gray!40!white,,text depth=-0.2mm]
\tikzstyle{red dot}=[gray dot] % for backwards-compatibility

\tikzstyle{black ddot}=[ddot,fill=black]
\tikzstyle{white ddot}=[ddot,fill=white]
\tikzstyle{gray ddot}=[ddot,fill=gray!40!white]

\tikzstyle{gray edge}=[gray!40!white]

\tikzstyle{small dot}=[inner sep=0.5mm,minimum width=0pt,minimum height=0pt,draw,shape=circle]

\tikzstyle{small black dot}=[small dot,fill=black]
\tikzstyle{small white dot}=[small dot,fill=white]
\tikzstyle{small gray dot}=[small dot,fill=gray!40!white]

\tikzstyle{causal dot}=[inner sep=0.4mm,minimum width=0pt,minimum height=0pt,draw=white,shape=circle,fill=gray!40!white]

\tikzstyle{phase dimensions}=[minimum size=5mm,font=\footnotesize,rectangle,rounded corners=2.5mm,inner sep=0.2mm,outer sep=-2mm]
\tikzstyle{dphase dimensions}=[minimum size=5mm,font=\footnotesize,rectangle,rounded corners=2.5mm,inner sep=0.2mm,outer sep=-2mm]
\tikzstyle{white phase dot}=[dot,fill=white,phase dimensions]
\tikzstyle{white phase ddot}=[ddot,fill=white,dphase dimensions]

\tikzstyle{white rect ddot}=[draw=black,fill=white,doubled,minimum size=5mm,font=\footnotesize,rectangle,rounded corners=2.5mm,inner sep=0.2mm]
\tikzstyle{gray rect ddot}=[draw=black,fill=gray!40!white,doubled,minimum size=6mm,font=\footnotesize,rectangle,rounded corners=3mm]

\tikzstyle{gray phase dot}=[dot,fill=gray!40!white,phase dimensions]
\tikzstyle{gray phase ddot}=[ddot,fill=gray!40!white,dphase dimensions]
\tikzstyle{grey phase dot}=[gray phase dot]
\tikzstyle{grey phase ddot}=[gray phase ddot]

\tikzstyle{small phase dimensions}=[minimum size=4mm,font=\tiny,rectangle,rounded corners=2mm,inner sep=0.2mm,outer sep=-2mm]
\tikzstyle{small dphase dimensions}=[minimum size=4mm,font=\tiny,rectangle,rounded corners=2mm,inner sep=0.2mm,outer sep=-2mm]

\tikzstyle{small gray phase dot}=[dot,fill=gray!40!white,small phase dimensions]
\tikzstyle{small gray phase ddot}=[ddot,fill=gray!40!white,small dphase dimensions]

\tikzstyle{small map}=[draw,shape=rectangle,minimum height=4mm,minimum width=4mm,fill=white]

\tikzstyle{cnot}=[fill=white,shape=circle,inner sep=-1.4pt]

\tikzstyle{asym hadamard}=[fill=white,draw,shape=NEbox,inner sep=0.6mm,font=\footnotesize,minimum height=4mm]
\tikzstyle{asym hadamard conj}=[fill=white,draw,shape=NWbox,inner sep=0.6mm,font=\footnotesize,minimum height=4mm]
\tikzstyle{asym hadamard dag}=[fill=white,draw,shape=SEbox,inner sep=0.6mm,font=\footnotesize,minimum height=4mm]

\tikzstyle{hadamard}=[fill=white,draw,inner sep=0.6mm,font=\footnotesize,minimum height=4mm,minimum width=4mm]
\tikzstyle{small hadamard}=[fill=white,draw,inner sep=0.6mm,minimum height=1.5mm,minimum width=1.5mm]
\tikzstyle{dhadamard}=[hadamard,doubled]
\tikzstyle{small dhadamard}=[small hadamard,doubled]
\tikzstyle{small dhadamard rotate}=[small hadamard,doubled,rotate=45]
\tikzstyle{antipode}=[white dot,inner sep=0.3mm,font=\footnotesize]

\tikzstyle{scalar}=[diamond,draw,inner sep=0.5pt,font=\small]
\tikzstyle{dscalar}=[diamond,doubled, draw,inner sep=0.5pt,font=\small]

\tikzstyle{small box}=[rectangle,inline text,fill=white,draw,minimum height=5mm,yshift=-0.5mm,minimum width=5mm,font=\small]
\tikzstyle{small gray box}=[small box,fill=gray!30]
\tikzstyle{medium box}=[rectangle,inline text,fill=white,draw,minimum height=5mm,yshift=-0.5mm,minimum width=10mm,font=\small]
\tikzstyle{square box}=[small box] % for backwards-compatibility
\tikzstyle{medium gray box}=[small box,fill=gray!30]
\tikzstyle{semilarge box}=[rectangle,inline text,fill=white,draw,minimum height=5mm,yshift=-0.5mm,minimum width=12.5mm,font=\small]
\tikzstyle{large box}=[rectangle,inline text,fill=white,draw,minimum height=5mm,yshift=-0.5mm,minimum width=15mm,font=\small]
\tikzstyle{large gray box}=[small box,fill=gray!30]

\tikzstyle{Bayes box}=[rectangle,fill=black,draw, minimum height=3mm, minimum width=3mm]

\tikzstyle{gray square point}=[small box,fill=gray!50]

\tikzstyle{dphase box white}=[dhadamard]
\tikzstyle{dphase box gray}=[dhadamard,fill=gray!50!white]

\tikzstyle{point}=[regular polygon,regular polygon sides=3,draw,scale=0.75,inner sep=-0.5pt,minimum width=9mm,fill=white,regular polygon rotate=180]
\tikzstyle{copoint}=[regular polygon,regular polygon sides=3,draw,scale=0.75,inner sep=-0.5pt,minimum width=9mm,fill=white]
\tikzstyle{dpoint}=[point,doubled]
\tikzstyle{dcopoint}=[copoint,doubled]

\tikzstyle{wide copoint}=[fill=white,draw,shape=isosceles triangle,shape border rotate=90,isosceles triangle stretches=true,inner sep=0pt,minimum width=1.5cm,minimum height=6.12mm]
\tikzstyle{wide point}=[fill=white,draw,shape=isosceles triangle,shape border rotate=-90,isosceles triangle stretches=true,inner sep=0pt,minimum width=1.5cm,minimum height=6.12mm,yshift=-0.0mm]
\tikzstyle{wide point plus}=[fill=white,draw,shape=isosceles triangle,shape border rotate=-90,isosceles triangle stretches=true,inner sep=0pt,minimum width=1.74cm,minimum height=7mm,yshift=-0.0mm]

\tikzstyle{wide dpoint}=[fill=white,doubled,draw,shape=isosceles triangle,shape border rotate=-90,isosceles triangle stretches=true,inner sep=0pt,minimum width=1.5cm,minimum height=6.12mm,yshift=-0.0mm]
\tikzstyle{wide dcopoint}=[fill=white,doubled,draw,shape=isosceles triangle,shape border rotate=90,isosceles triangle stretches=true,inner sep=0pt,minimum width=1.5cm,minimum height=6.12mm,yshift=-0.0mm]

\tikzstyle{tinypoint}=[regular polygon,regular polygon sides=3,draw,scale=0.55,inner sep=-0.15pt,minimum width=6mm,fill=white,regular polygon rotate=180]

\tikzstyle{white point}=[point]
\tikzstyle{white dpoint}=[dpoint]
\tikzstyle{green point}=[white point] % for backwards-compatibility
\tikzstyle{white copoint}=[copoint]
\tikzstyle{gray point}=[point,fill=gray!40!white]
\tikzstyle{gray dpoint}=[gray point,doubled]
\tikzstyle{red point}=[gray point] % for backwards-compatibility
\tikzstyle{gray copoint}=[copoint,fill=gray!40!white]
\tikzstyle{gray dcopoint}=[gray copoint,doubled]

\tikzstyle{white point guide}=[regular polygon,regular polygon sides=3,font=\scriptsize,draw,scale=0.65,inner sep=-0.5pt,minimum width=9mm,fill=white,regular polygon rotate=180]

\tikzstyle{black point}=[point,fill=black,font=\color{white}]
\tikzstyle{black copoint}=[copoint,fill=black,font=\color{white}]

\tikzstyle{tiny gray point}=[tinypoint,fill=gray!40!white]

\tikzstyle{diredge}=[->]
\tikzstyle{ddiredge}=[<->]
\tikzstyle{rdiredge}=[<-]
\tikzstyle{thickdiredge}=[->, very thick]
\tikzstyle{pointer edge}=[->,very thick,gray]
\tikzstyle{pointer edge part}=[very thick,gray]
\tikzstyle{dashed edge}=[dashed]
\tikzstyle{thick dashed edge}=[very thick,dashed]
\tikzstyle{thick gray dashed edge}=[thick dashed edge,gray!40]
\tikzstyle{thick map edge}=[very thick,|->]

\makeatletter
\newcommand{\boxshape}[3]{%
\pgfdeclareshape{#1}{
\inheritsavedanchors[from=rectangle] % this is nearly a rectangle
\inheritanchorborder[from=rectangle]
\inheritanchor[from=rectangle]{center}
\inheritanchor[from=rectangle]{north}
\inheritanchor[from=rectangle]{south}
\inheritanchor[from=rectangle]{west}
\inheritanchor[from=rectangle]{east}
\backgroundpath{% this is new
\southwest \pgf@xa=\pgf@x \pgf@ya=\pgf@y
\northeast \pgf@xb=\pgf@x \pgf@yb=\pgf@y

\@tempdima=#2
\@tempdimb=#3

\pgfpathmoveto{\pgfpoint{\pgf@xa - 5pt + \@tempdima}{\pgf@ya}}
\pgfpathlineto{\pgfpoint{\pgf@xa - 5pt - \@tempdima}{\pgf@yb}}
\pgfpathlineto{\pgfpoint{\pgf@xb + 5pt + \@tempdimb}{\pgf@yb}}
\pgfpathlineto{\pgfpoint{\pgf@xb + 5pt - \@tempdimb}{\pgf@ya}}
\pgfpathlineto{\pgfpoint{\pgf@xa - 5pt + \@tempdima}{\pgf@ya}}
\pgfpathclose
}
}}

\boxshape{NEbox}{0pt}{5pt}
\boxshape{SEbox}{0pt}{-5pt}
\boxshape{NWbox}{5pt}{0pt}
\boxshape{SWbox}{-5pt}{0pt}
\boxshape{EBox}{-3pt}{3pt}
\boxshape{WBox}{3pt}{-3pt}
\makeatother

\tikzstyle{cloud}=[shape=cloud,draw,minimum width=1.5cm,minimum height=1.5cm]

\tikzstyle{map}=[draw,shape=NEbox,inner sep=2pt,minimum height=6mm,fill=white]
\tikzstyle{dashedmap}=[draw,dashed,shape=NEbox,inner sep=2pt,minimum height=6mm,fill=white]
\tikzstyle{mapdag}=[draw,shape=SEbox,inner sep=2pt,minimum height=6mm,fill=white]
\tikzstyle{mapadj}=[draw,shape=SEbox,inner sep=2pt,minimum height=6mm,fill=white]
\tikzstyle{maptrans}=[draw,shape=SWbox,inner sep=2pt,minimum height=6mm,fill=white]
\tikzstyle{mapconj}=[draw,shape=NWbox,inner sep=2pt,minimum height=6mm,fill=white]

\tikzstyle{medium map}=[draw,shape=NEbox,inner sep=2pt,minimum height=6mm,fill=white,minimum width=7mm]
\tikzstyle{medium map dag}=[draw,shape=SEbox,inner sep=2pt,minimum height=6mm,fill=white,minimum width=7mm]
\tikzstyle{medium map adj}=[draw,shape=SEbox,inner sep=2pt,minimum height=6mm,fill=white,minimum width=7mm]
\tikzstyle{medium map trans}=[draw,shape=SWbox,inner sep=2pt,minimum height=6mm,fill=white,minimum width=7mm]
\tikzstyle{medium map conj}=[draw,shape=NWbox,inner sep=2pt,minimum height=6mm,fill=white,minimum width=7mm]
\tikzstyle{semilarge map}=[draw,shape=NEbox,inner sep=2pt,minimum height=6mm,fill=white,minimum width=9.5mm]
\tikzstyle{semilarge map trans}=[draw,shape=SWbox,inner sep=2pt,minimum height=6mm,fill=white,minimum width=9.5mm]
\tikzstyle{semilarge map adj}=[draw,shape=SEbox,inner sep=2pt,minimum height=6mm,fill=white,minimum width=9.5mm]
\tikzstyle{semilarge map dag}=[draw,shape=SEbox,inner sep=2pt,minimum height=6mm,fill=white,minimum width=9.5mm]
\tikzstyle{semilarge map conj}=[draw,shape=NWbox,inner sep=2pt,minimum height=6mm,fill=white,minimum width=9.5mm]
\tikzstyle{large map}=[draw,shape=NEbox,inner sep=2pt,minimum height=6mm,fill=white,minimum width=12mm]
\tikzstyle{large map conj}=[draw,shape=NWbox,inner sep=2pt,minimum height=6mm,fill=white,minimum width=12mm]
\tikzstyle{very large map}=[draw,shape=NEbox,inner sep=2pt,minimum height=6mm,fill=white,minimum width=17mm]

\tikzstyle{medium dmap}=[draw,doubled,shape=NEbox,inner sep=2pt,minimum height=6mm,fill=white,minimum width=7mm]
\tikzstyle{medium dmap dag}=[draw,doubled,shape=SEbox,inner sep=2pt,minimum height=6mm,fill=white,minimum width=7mm]
\tikzstyle{medium dmap adj}=[draw,doubled,shape=SEbox,inner sep=2pt,minimum height=6mm,fill=white,minimum width=7mm]
\tikzstyle{medium dmap trans}=[draw,doubled,shape=SWbox,inner sep=2pt,minimum height=6mm,fill=white,minimum width=7mm]
\tikzstyle{medium dmap conj}=[draw,doubled,shape=NWbox,inner sep=2pt,minimum height=6mm,fill=white,minimum width=7mm]
\tikzstyle{semilarge dmap}=[draw,doubled,shape=NEbox,inner sep=2pt,minimum height=6mm,fill=white,minimum width=9.5mm]
\tikzstyle{semilarge dmap trans}=[draw,doubled,shape=SWbox,inner sep=2pt,minimum height=6mm,fill=white,minimum width=9.5mm]
\tikzstyle{semilarge dmap adj}=[draw,doubled,shape=SEbox,inner sep=2pt,minimum height=6mm,fill=white,minimum width=9.5mm]
\tikzstyle{semilarge dmap dag}=[draw,doubled,shape=SEbox,inner sep=2pt,minimum height=6mm,fill=white,minimum width=9.5mm]
\tikzstyle{semilarge dmap conj}=[draw,doubled,shape=NWbox,inner sep=2pt,minimum height=6mm,fill=white,minimum width=9.5mm]
\tikzstyle{large dmap}=[draw,doubled,shape=NEbox,inner sep=2pt,minimum height=6mm,fill=white,minimum width=12mm]
\tikzstyle{large dmap conj}=[draw,doubled,shape=NWbox,inner sep=2pt,minimum height=6mm,fill=white,minimum width=12mm]
\tikzstyle{large dmap trans}=[draw,doubled,shape=SWbox,inner sep=2pt,minimum height=6mm,fill=white,minimum width=12mm]
\tikzstyle{large dmap adj}=[draw,doubled,shape=SEbox,inner sep=2pt,minimum height=6mm,fill=white,minimum width=12mm]
\tikzstyle{large dmap dag}=[draw,doubled,shape=SEbox,inner sep=2pt,minimum height=6mm,fill=white,minimum width=12mm]
\tikzstyle{very large dmap}=[draw,doubled,shape=NEbox,inner sep=2pt,minimum height=6mm,fill=white,minimum width=19.5mm]

\tikzstyle{muxbox}=[draw,shape=rectangle,minimum height=3mm,minimum width=3mm,fill=white]
\tikzstyle{dmuxbox}=[muxbox,doubled]

\tikzstyle{box}=[draw,shape=rectangle,inner sep=2pt,minimum height=6mm,minimum width=6mm,fill=white]
\tikzstyle{dbox}=[draw,doubled,shape=rectangle,inner sep=2pt,minimum height=6mm,minimum width=6mm,fill=white]
\tikzstyle{dmap}=[draw,doubled,shape=NEbox,inner sep=2pt,minimum height=6mm,fill=white]
\tikzstyle{dmapdag}=[draw,doubled,shape=SEbox,inner sep=2pt,minimum height=6mm,fill=white]
\tikzstyle{dmapadj}=[draw,doubled,shape=SEbox,inner sep=2pt,minimum height=6mm,fill=white]
\tikzstyle{dmaptrans}=[draw,doubled,shape=SWbox,inner sep=2pt,minimum height=6mm,fill=white]
\tikzstyle{dmapconj}=[draw,doubled,shape=NWbox,inner sep=2pt,minimum height=6mm,fill=white]

\tikzstyle{ddmap}=[draw,doubled,dashed,shape=NEbox,inner sep=2pt,minimum height=6mm,fill=white]
\tikzstyle{ddmapdag}=[draw,doubled,dashed,shape=SEbox,inner sep=2pt,minimum height=6mm,fill=white]
\tikzstyle{ddmapadj}=[draw,doubled,dashed,shape=SEbox,inner sep=2pt,minimum height=6mm,fill=white]
\tikzstyle{ddmaptrans}=[draw,doubled,dashed,shape=SWbox,inner sep=2pt,minimum height=6mm,fill=white]
\tikzstyle{ddmapconj}=[draw,doubled,dashed,shape=NWbox,inner sep=2pt,minimum height=6mm,fill=white]

\boxshape{sNEbox}{0pt}{3pt}
\boxshape{sSEbox}{0pt}{-3pt}
\boxshape{sNWbox}{3pt}{0pt}
\boxshape{sSWbox}{-3pt}{0pt}
\tikzstyle{smap}=[draw,shape=sNEbox,fill=white]
\tikzstyle{smapdag}=[draw,shape=sSEbox,fill=white]
\tikzstyle{smapadj}=[draw,shape=sSEbox,fill=white]
\tikzstyle{smaptrans}=[draw,shape=sSWbox,fill=white]
\tikzstyle{smapconj}=[draw,shape=sNWbox,fill=white]

\tikzstyle{dsmap}=[draw,dashed,shape=sNEbox,fill=white]
\tikzstyle{dsmapdag}=[draw,dashed,shape=sSEbox,fill=white]
\tikzstyle{dsmaptrans}=[draw,dashed,shape=sSWbox,fill=white]
\tikzstyle{dsmapconj}=[draw,dashed,shape=sNWbox,fill=white]

\boxshape{mNEbox}{0pt}{10pt}
\boxshape{mSEbox}{0pt}{-10pt}
\boxshape{mNWbox}{10pt}{0pt}
\boxshape{mSWbox}{-10pt}{0pt}
\tikzstyle{mmap}=[draw,shape=mNEbox]
\tikzstyle{mmapdag}=[draw,shape=mSEbox]
\tikzstyle{mmaptrans}=[draw,shape=mSWbox]
\tikzstyle{mmapconj}=[draw,shape=mNWbox]

\tikzstyle{mmapgray}=[draw,fill=gray!40!white,shape=mNEbox]
\tikzstyle{smapgray}=[draw,fill=gray!40!white,shape=sNEbox]

\makeatletter
\pgfdeclareshape{cornerpoint}{
\inheritsavedanchors[from=rectangle] % this is nearly a rectangle
\inheritanchorborder[from=rectangle]
\inheritanchor[from=rectangle]{center}
\inheritanchor[from=rectangle]{north}
\inheritanchor[from=rectangle]{south}
\inheritanchor[from=rectangle]{west}
\inheritanchor[from=rectangle]{east}
\backgroundpath{% this is new
\southwest \pgf@xa=\pgf@x \pgf@ya=\pgf@y
\northeast \pgf@xb=\pgf@x \pgf@yb=\pgf@y

\pgfmathsetmacro{\pgf@shorten@left}{\pgfkeysvalueof{/tikz/shorten left}}
\pgfmathsetmacro{\pgf@shorten@right}{\pgfkeysvalueof{/tikz/shorten right}}

\pgfpathmoveto{\pgfpoint{0.5 * (\pgf@xa + \pgf@xb)}{\pgf@ya - 5pt}}
\pgfpathlineto{\pgfpoint{\pgf@xa - 8pt + \pgf@shorten@left}{\pgf@yb - 1.5 * \pgf@shorten@left}}
\pgfpathlineto{\pgfpoint{\pgf@xa - 8pt + \pgf@shorten@left}{\pgf@yb}}
\pgfpathlineto{\pgfpoint{\pgf@xb + 8pt - \pgf@shorten@right}{\pgf@yb}}
\pgfpathlineto{\pgfpoint{\pgf@xb + 8pt - \pgf@shorten@right}{\pgf@yb - 1.5 * \pgf@shorten@right}}
\pgfpathclose
}
}

\pgfdeclareshape{cornercopoint}{
\inheritsavedanchors[from=rectangle] % this is nearly a rectangle
\inheritanchorborder[from=rectangle]
\inheritanchor[from=rectangle]{center}
\inheritanchor[from=rectangle]{north}
\inheritanchor[from=rectangle]{south}
\inheritanchor[from=rectangle]{west}
\inheritanchor[from=rectangle]{east}
\backgroundpath{% this is new
\southwest \pgf@xa=\pgf@x \pgf@ya=\pgf@y
\northeast \pgf@xb=\pgf@x \pgf@yb=\pgf@y

\pgfmathsetmacro{\pgf@shorten@left}{\pgfkeysvalueof{/tikz/shorten left}}
\pgfmathsetmacro{\pgf@shorten@right}{\pgfkeysvalueof{/tikz/shorten right}}

\pgfpathmoveto{\pgfpoint{0.5 * (\pgf@xa + \pgf@xb)}{\pgf@yb + 5pt}}
\pgfpathlineto{\pgfpoint{\pgf@xa - 8pt + \pgf@shorten@left}{\pgf@ya + 1.5 * \pgf@shorten@left}}
\pgfpathlineto{\pgfpoint{\pgf@xa - 8pt + \pgf@shorten@left}{\pgf@ya}}
\pgfpathlineto{\pgfpoint{\pgf@xb + 8pt - \pgf@shorten@right}{\pgf@ya}}
\pgfpathlineto{\pgfpoint{\pgf@xb + 8pt - \pgf@shorten@right}{\pgf@ya + 1.5 * \pgf@shorten@right}}
\pgfpathclose
}
}

\pgfdeclareshape{langpoint}{
\inheritsavedanchors[from=rectangle] % this is nearly a rectangle
\inheritanchorborder[from=rectangle]
\inheritanchor[from=rectangle]{center}
\inheritanchor[from=rectangle]{north}
\inheritanchor[from=rectangle]{south}
\inheritanchor[from=rectangle]{west}
\inheritanchor[from=rectangle]{east}
\backgroundpath{% this is new
\southwest \pgf@xa=\pgf@x \pgf@ya=\pgf@y
\northeast \pgf@xb=\pgf@x \pgf@yb=\pgf@y

\pgfmathsetmacro{\pgf@shorten@left}{\pgfkeysvalueof{/tikz/shorten left}}
\pgfmathsetmacro{\pgf@shorten@right}{\pgfkeysvalueof{/tikz/shorten right}}

\pgfpathmoveto{\pgfpoint{0.5 * (\pgf@xa + \pgf@xb)}{\pgf@ya - 2pt}}
\pgfpathlineto{\pgfpoint{\pgf@xa - 8pt}{\pgf@yb - 3 * \pgf@shorten@left + 5pt}} 
\pgfpathlineto{\pgfpoint{\pgf@xa - 8pt}{\pgf@yb -1pt}}
\pgfpathlineto{\pgfpoint{\pgf@xb + 8pt}{\pgf@yb -1pt}}
\pgfpathlineto{\pgfpoint{\pgf@xb + 8pt}{\pgf@yb - 3 * \pgf@shorten@left + 5pt}}%NEW POINT
\pgfpathclose
}
}

\pgfdeclareshape{langcopoint}{
\inheritsavedanchors[from=rectangle] % this is nearly a rectangle
\inheritanchorborder[from=rectangle]
\inheritanchor[from=rectangle]{center}
\inheritanchor[from=rectangle]{north}
\inheritanchor[from=rectangle]{south}
\inheritanchor[from=rectangle]{west}
\inheritanchor[from=rectangle]{east}
\backgroundpath{% this is new
\southwest \pgf@xa=\pgf@x \pgf@ya=\pgf@y
\northeast \pgf@xb=\pgf@x \pgf@yb=\pgf@y

\pgfmathsetmacro{\pgf@shorten@left}{\pgfkeysvalueof{/tikz/shorten left}}
\pgfmathsetmacro{\pgf@shorten@right}{\pgfkeysvalueof{/tikz/shorten right}}

\pgfpathmoveto{\pgfpoint{0.5 * (\pgf@xa + \pgf@xb)}{\pgf@yb +0pt}}
\pgfpathlineto{\pgfpoint{\pgf@xa - 8pt}{\pgf@ya + 3 * \pgf@shorten@left - 5pt}} 
\pgfpathlineto{\pgfpoint{\pgf@xa - 8pt}{\pgf@ya + 1pt}}
\pgfpathlineto{\pgfpoint{\pgf@xb + 8pt}{\pgf@ya + 1pt}}
\pgfpathlineto{\pgfpoint{\pgf@xb + 8pt}{\pgf@ya + 3 * \pgf@shorten@left - 5pt}}%NEW POINT
\pgfpathclose
}
}

\pgfdeclareshape{langrect}{
\inheritsavedanchors[from=rectangle] % this is nearly a rectangle
\inheritanchorborder[from=rectangle]
\inheritanchor[from=rectangle]{center}
\inheritanchor[from=rectangle]{north}
\inheritanchor[from=rectangle]{south}
\inheritanchor[from=rectangle]{west}
\inheritanchor[from=rectangle]{east}
\backgroundpath{% this is new
\southwest \pgf@xa=\pgf@x \pgf@ya=\pgf@y
\northeast \pgf@xb=\pgf@x \pgf@yb=\pgf@y

\pgfmathsetmacro{\pgf@shorten@left}{\pgfkeysvalueof{/tikz/shorten left}}
\pgfmathsetmacro{\pgf@shorten@right}{\pgfkeysvalueof{/tikz/shorten right}}

\pgfpathmoveto{\pgfpoint{\pgf@xa - 8pt}{\pgf@ya + 3 * \pgf@shorten@left - 5pt}} 
\pgfpathlineto{\pgfpoint{\pgf@xa - 8pt}{\pgf@ya + 1pt}}
\pgfpathlineto{\pgfpoint{\pgf@xb + 8pt}{\pgf@ya + 1pt}}
\pgfpathlineto{\pgfpoint{\pgf@xb + 8pt}{\pgf@ya + 3 * \pgf@shorten@left - 5pt}}%NEW POINT
\pgfpathclose
}
}
\makeatother

\pgfkeyssetvalue{/tikz/shorten left}{0pt}
\pgfkeyssetvalue{/tikz/shorten right}{0pt}

\tikzstyle{kpoint common}=[draw,fill=white,inner sep=1pt,minimum height=4mm]

\tikzstyle{langstate}=[shape=langcopoint,shorten left=5pt,kpoint common,font=\footnotesize]
\tikzstyle{langeffect}=[shape=langpoint,shorten left=5pt,kpoint common,font=\footnotesize]
\tikzstyle{langbox}=[shape=langrect,shorten left=5pt,kpoint common,font=\footnotesize] 
\tikzstyle{kpoint}=[shape=cornerpoint,shorten left=5pt,kpoint common]
\tikzstyle{kpoint adjoint}=[shape=cornercopoint,shorten left=5pt,kpoint common]

\tikzstyle{kpoint conjugate}=[shape=cornerpoint,shorten right=5pt,kpoint common]
\tikzstyle{kpoint transpose}=[shape=cornercopoint,shorten right=5pt,kpoint common]
\tikzstyle{kpoint symm}=[shape=cornerpoint,shorten left=5pt,shorten right=5pt,kpoint common]

\tikzstyle{black kpoint}=[shape=cornerpoint,shorten left=5pt,kpoint common,fill=black,font=\color{white}]
\tikzstyle{black kpoint adjoint}=[shape=cornercopoint,shorten left=5pt,kpoint common,fill=black,font=\color{white}]
\tikzstyle{black kpointadj}=[shape=cornercopoint,shorten left=5pt,kpoint common,fill=black,font=\color{white}]

\tikzstyle{black dkpoint}=[shape=cornerpoint,shorten left=5pt,kpoint common,fill=black, doubled,font=\color{white}]
\tikzstyle{black dkpoint adjoint}=[shape=cornercopoint,shorten left=5pt,kpoint common,fill=black, doubled,font=\color{white}]
\tikzstyle{black dkpointadj}=[shape=cornercopoint,shorten left=5pt,kpoint common,fill=black, doubled,font=\color{white}]

\tikzstyle{kpointdag}=[kpoint adjoint]
\tikzstyle{kpointadj}=[kpoint adjoint]
\tikzstyle{kpointconj}=[kpoint conjugate]
\tikzstyle{kpointtrans}=[kpoint transpose]

\tikzstyle{big kpoint}=[kpoint, minimum width=1.2 cm, minimum height=8mm, inner sep=4pt, text depth=3mm]

\tikzstyle{wide kpoint}=[kpoint, minimum width=1 cm, inner sep=2pt]%, text depth=-0.7 mm]
\tikzstyle{wide kpointdag}=[kpointdag, minimum width=1 cm, inner sep=2pt]%, text depth=0.7 mm]
\tikzstyle{wide kpointconj}=[kpointconj, minimum width=1 cm, inner sep=2pt]%, text depth=-0.7 mm]
\tikzstyle{wide kpointtrans}=[kpointtrans, minimum width=1 cm, inner sep=2pt]%, text depth=0.7 mm]

\tikzstyle{gray kpoint}=[kpoint,fill=gray!50!white]
\tikzstyle{gray kpointdag}=[kpointdag,fill=gray!50!white]
\tikzstyle{gray kpointadj}=[kpointadj,fill=gray!50!white]
\tikzstyle{gray kpointconj}=[kpointconj,fill=gray!50!white]
\tikzstyle{gray kpointtrans}=[kpointtrans,fill=gray!50!white]

\tikzstyle{gray dkpoint}=[kpoint,fill=gray!50!white,doubled]
\tikzstyle{gray dkpointdag}=[kpointdag,fill=gray!50!white,doubled]
\tikzstyle{gray dkpointadj}=[kpointadj,fill=gray!50!white,doubled]
\tikzstyle{gray dkpointconj}=[kpointconj,fill=gray!50!white,doubled]
\tikzstyle{gray dkpointtrans}=[kpointtrans,fill=gray!50!white,doubled]

\tikzstyle{white label}=[draw,fill=white,rectangle,inner sep=0.7 mm]
\tikzstyle{gray label}=[draw,fill=gray!50!white,rectangle,inner sep=0.7 mm]
\tikzstyle{black label}=[draw,fill=black,rectangle,inner sep=0.7 mm]

\tikzstyle{dkpoint}=[kpoint,doubled]
\tikzstyle{wide dkpoint}=[wide kpoint,doubled]
\tikzstyle{dkpointdag}=[kpoint adjoint,doubled]
\tikzstyle{wide dkpointdag}=[wide kpointdag,doubled]
\tikzstyle{dkcopoint}=[kpoint adjoint,doubled]
\tikzstyle{dkpointadj}=[kpoint adjoint,doubled]
\tikzstyle{dkpointconj}=[kpoint conjugate,doubled]
\tikzstyle{dkpointtrans}=[kpoint transpose,doubled]

\tikzstyle{kscalar}=[kpoint common, shape=EBox, inner xsep=-1pt, inner ysep=3pt,font=\small]
\tikzstyle{kscalarconj}=[kpoint common, shape=WBox, inner xsep=-1pt, inner ysep=3pt,font=\small]

 \tikzstyle{upground}=[circuit ee IEC,ground,rotate=90,scale=2.5]
 \tikzstyle{downground}=[circuit ee IEC,ground,rotate=-90,scale=2.5]
 \tikzstyle{bigground}=[regular polygon,regular polygon sides=3,draw=gray,scale=0.50,inner sep=-0.5pt,minimum width=10mm,fill=gray]
\tikzstyle{arrs}=[-latex,font=\small,auto]
\tikzstyle{arrow plain}=[arrs]
\tikzstyle{arrow dashed}=[dashed,arrs]
\tikzstyle{arrow bold}=[very thick,arrs]
\tikzstyle{arrow hide}=[draw=white!0,-]
\tikzstyle{arrow reverse}=[latex-]
\tikzstyle{cdnode}=[]

\definecolor{hexcolor0xa9a9a9}{rgb}{0.663,0.663,0.663}
\tikzstyle{GrayLine}=[dashed,draw=hexcolor0xa9a9a9] 
\tikzstyle{gray}=[dashed,draw=hexcolor0xa9a9a9]

\def\bR{\begin{color}{red}}  
\def\bB{\begin{color}{blue}}
\def\bM{\begin{color}{magenta}} 
\def\bC{\begin{color}{cyan}}
\def\bW{\begin{color}{white}}
\def\bBl{\begin{color}{black}}
\def\bG{\begin{color}{green}}
\def\bY{\begin{color}{yellow}}
\def\e{\end{color}\xspace}
\newcommand{\bit}{\begin{itemize}}
\newcommand{\eit}{\end{itemize}\par\noindent}
\newcommand{\ben}{\begin{enumerate}}
\newcommand{\een}{\end{enumerate}\par\noindent}
\newcommand{\beq}{\begin{equation}}
\newcommand{\eeq}{\end{equation}\par\noindent}
\newcommand{\beqa}{\begin{eqnarray*}}
\newcommand{\eeqa}{\end{eqnarray*}\par\noindent}
\newcommand{\beqn}{\begin{eqnarray}}
\newcommand{\eeqn}{\end{eqnarray}\par\noindent}

\title*{The Mathematics of Text Structure} %Sentence Composition}

\author{Bob Coecke}
\institute{Bob Coecke \at 
Oxford University, Department of Computer Science\\ 
Cambridge Quantum Computing Ltd.\\ \email{coecke@cs.ox.ac.uk, bob.coecke@cambridgequantum.com}}

\begin{document}  
\maketitle 

%%% Springer %%%
\abstract{ 
%%% BOB %%%\begin{abstract}  
In previous work we gave a mathematical foundation, referred to as DisCoCat, for how words interact in a sentence in order to produce the meaning of that sentence. To do so, we exploited the perfect structural match of grammar and categories of meaning spaces.  
\newline\indent
Here, we give a mathematical foundation, referred to as DisCoCirc, for how sentences interact in texts in order to produce the meaning of that text. First we revisit DisCoCat. While in DisCoCat  all meanings are fixed as states (i.e.~have no input), in  DisCoCirc word meanings correspond to a type, or system, and the states  of this system can evolve.  Sentences are gates within a circuit which update the variable meanings of those words.  
\newline\indent 
Like in DisCoCat, word meanings  can live in a variety of  spaces e.g.~propositional, vectorial, or cognitive.  The compositional structure are string diagrams representing information flows,
 %induced by grammar and functional words.  
and  an entire text yields a single string diagram in which word meanings lift to the meaning of the entire text.
\newline\indent 
While the developments in this paper are independent of a physical embodiment (cf.~classical vs.~quantum computing), both the compositional formalism and suggested meaning model are highly quantum-inspired, and implementation on a quantum computer would come with a range of benefits.
\newline\indent
We also praise Jim Lambek for his role in mathematical linguistics in general, and  the development of   the DisCo program more specifically. 
} 
%%% BOB %%%\end{abstract}    
   
%%% BOB %%%
%\tableofcontents
 
\section{Introduction}\label{sec:intro}     

DisCoCat (cf.~\underline{Cat}egorical \underline{Co}mpositional \underline{Dis}tributional) \cite{CSC} resulted from addressing the following question: 
\begin{center}
{\it There are dictionaries for words.\\ Why aren't there any dictionaries for sentences?}  
\end{center}
This question is both of academic interest as well as of practical interest.  Firstly, it addresses how we can understand sentences that we never heard before, provided we understand the words.  Secondly, it could enable machines to do so too---see e.g.~\cite{ClarkPulman} for a preliminary discussion.  

Now, one obviously can extend the question to: 
\begin{center}
{\it Why aren't there any dictionaries for entire texts?}  
\end{center}
While there is no such thing like a grammar that very rigidly organises sentences in a text, there should be something else structuring  sentences in a text, as, to put it naively, we can't just swap sentences randomly around while retaining the meaning of the text. What exactly is the structure  governing sentences? 
Also, how does it relate to the structure governing words in sentences?   
Overall, this paper has two purposes:
\bit
\item To give a mathematical foundation for  sentence meaning composition, and how sentence meaning interacts with word meaning composition, by  modifying as well as elaborating upon our theory of grammar-driven word meaning composition known as DisCoCat. 
\item As this paper was commissioned for a volume dedicated to the great Jim Lambek, to praise Lambek for the development of mathematical linguistics in general,  for his role in the development of   the DisCoCat program more specifically, and for the role of  physics in all of that \cite{teleling}.
\eit
Concerning the 1st goal, we believe that the results in this paper may constitute radical progress for the broad DisCoCat research program, as three important longstanding issues are addressed `in full generality':
\ben
\item[a.]  How   \underline{sentence meanings compose} in order to form the meaning of  text. 
\item[b.] How \underline{word meanings evolve} in text, when learning new things.
\item[c.] What the type of the \underline{sentence meaning space} is. 
\een
The key idea to achieve (a) is to not treat (all) word-meanings as static entities, like we did in DisCoCat, but as dynamic ones, so that they can evolve in the light of what is conveyed about that word within the text, i.e.~(b).  For example, using a movie analogy, we can learn more about the main actors of the text, and/or these actors themselves can learn about the rest of the world.  Technically, rather than states, such actors will be represented by types, sentences then act on those types as I/O-processes, transforming some initial state of an actor into a resulting one, which yields (c). In other words, text made up of several sentences is organised as a \em circuit\em.  Since: 
\ben
\item Word-meanings:\ \  states$\ \ \mapsto\ \ $types 
\item Sentence-meanings:\ \   states$\ \ \mapsto\ \ $I/O-processes  
\item Text-meaning:\ \   $\emptyset\ \ \mapsto\ \ $circuit
\een
are a radical departure from the state-focussed DisCoCat program, we name it DisCoCirc (cf.~\underline{Circ}uit-shaped \underline{Co}mpositional \underline{Dis}tributional). Other than that, all of the attractive features of DisCoCat are all retained (see Sec.~\ref{Sec:CatFromCirc}), such as sentences with different grammatical structure having the same type, model-flexibility, and  the diagrammatic format  (cf.~Sec.~\ref{sec:featflaw}). In fact, DisCoCat can be seen as an instance of DisCoCirc, in that in any DisCoCirc(uit) there will be multiple participating DisCoCat(s)---and the gloves are off too:%\TODOb{Improve pic \& scan.} %\TODOb{DisCoCirc a bit more circuit-like.}
%  \begin{center}
%\epsfig{figure=Cartoon,width=160pt}
%  \end{center}  
  \[
 \quad\ \,\begin{array}{ccc}
\raisebox{12mm}{\epsfig{figure=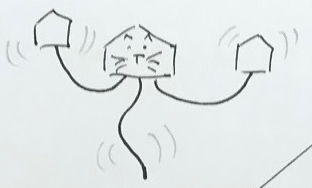,width=91pt}}&\qquad\quad \ \ \qquad&\epsfig{figure=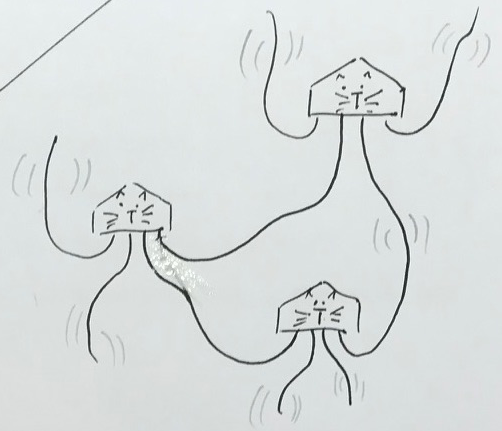,width=146pt}\qquad\\
\mbox{\rm DisCoCat} & &\mbox{\rm DisCoCirc}\quad\ \,
\end{array}
  \]

\par\medskip\noindent
{\bf Jim Lambek's legacy.} We already started our praise for Lambek in the title of this paper, by  tweaking the title of Lambek's seminal paper ``The Mathematics of Sentence Structure'' \cite{Lambek0}.   
%which put forward the now widely agreed mathematics of grammar, \em Lambek grammar\em. While this was done some 60 years ago,  
Not much has changed in that Lambek's story is pretty much still what mathematical linguists agree on today. Everyone?  Well, Lambek himself certainly did not.  Some 20 years ago Lambek  decided to `do a von Neumann',\footnote{von Neumann denounced his own quantum mechanical formalism merely three years after it was published as a book \cite{vN}, and devoted a large portion of the remainder of his life to finding an alternative formalism---see \cite{Redei1} for a discussion.} and replaced \em Lambek grammar \em with \em pregroup grammar \em \cite{Lambek1, LambekBook}. Around 2004 I was giving a talk at the McGill category theory seminar about our then new diagrammatic description of quantum teleportation \cite{AC1, Kindergarten}. Lambek immediately pointed out: ``Those are pre-groups!"  The compact closed category-theory underpinning of quantum theory, have indeed pregroups as the posetal instance.  It was this connection, between grammar and teleportation diagrams, that inspired  the DisCoCat model, making it look as if word-meaning being teleported around in sentences by means of the channels provided by the pregroup grammar \cite{teleling}.  Lambek himself explicitly stressed this connection between language and physics in a paper written in 2006 \cite{LambekLNP},  which due to my all too slow editorial skills only appeared in 2011.     

Lambek made many more pioneering contributions, including significant contributions to linear logic, which we briefly get into in Sec.~\ref{Sec:AND}, and of course there also is the Curry-Howard-Lambek isomorphism \cite{lambek1968deductive, lambek1969deductive, lambek1980lambda, lambek1993categorial}.  Just as in that context programming becomes an instance of category theory, something very similar is true in the case of DisCoCat for language. 

%Let me  point out maybe Lambek's greatest virtue: in an environment of people who took themselves way too serious, he was always up for a lighter note, and crack a joke.  Just to mention one, for our categorical underpinning of quantum theory we make use of something called a dagger-structure.  When Lambek gave a talk at Oxford in 2006 about connections between language and physics, he omitted daggers ``as he couldn't take those on the plane".  

%\bR Most importantly, the passage from Lambek grammar to pregroups is key for the passage to DisCoCirc, as it further exploits the intuitive presentation on how grammar mediates the interaction of word-meanings in sentences, now augmented to entire texts.\e
 
\par\medskip\noindent
{\bf The  need for structural understanding.}  Being from a pure mathematician, Lambek's work on the `real world' was heavily structurally driven (and so is ours in this paper).  These days prediction-driven work drawing from big data has clearly taken the forefront.  Allow us to share some reflections on that, by looking at some key historical scientific developments.

In particular, undeniably, Natural Language Processing (NLP) has made great progress thanks to the great progress recently made in Machine Learning (ML). On the other hand it is fair to say that this hasn't necessarily increased our structural understanding of language.  It would be a major mistake to only follow the path where empirical success takes us, and ignore that which increases understanding, as we have learned the hard way from how (the most important ever) progress has been made in physics. 

The discovery of the theory of Newtonian mechanics came from the study of movements of planets and stars.  This study was data driven as these movements are vital for ships on sea to find their way home. The longest surviving model was Ptolemy's epicycle model. While there were Copernicus, Kepler, Galilei and Newton,  it took until Einstein for that line of research  to match Ptolemy's correct predictions,  as the latter accounted for relativistic effects. The reason was simply that for any anomaly one could always add a sufficient number of  epicycles. This shows that while structural insights (cf.~Earth not in the middle of the universe), may take time to get the predictions right, but in the end even make the data driven models obsolete.\footnote{While neural networks have an ontological underpinning taking some inspiration from the human brain,  the  universal approximation theorem for neural networks \cite{cybenko1989approximations}, which states that one can approximate any continuous function on compact subsets of $\mathbb{R}^n$, seems to be somewhat on par with the unrestricted adding of epicycles.}  

There is a deeper aspect to this story, namely: Why did we have to go to outer space to see mechanics in action in a manner that we could truly \underline{understand} it?  The simple reason is that the role of friction on earth is so imposing that one was rather led to take on the Aristotelian point of view that any movement requires force. Only in outer space we see frictionless movement.   We think that the same is true for language: the manner it is used is full of deviations from the structural core.  We believe that just like in the case of Newtonian mechanics, understanding that structural core will ultimately lead to much better predictive power.  As an analogy, if ML would truly solve all of our problems, then one would expect it to be used to calculate trajectories of space ships and compute quantum spectra, learning from the available experimental data, but I personally don't expect that we will dump non-Euclidean geometry and Hilbert space for these purposes, despite the fact that real physics calculations, for many reasons, all deviate from the ideal.  That said, ML could probably have saved Copernicus a hell of a lot of time when analysing Brahe's data.
 
\par\medskip\noindent
{\bf Why pregroups?} Pregroups are easily the simplest model of grammar, and have a  very simple graphical presentation as wires-structures. We think that such a simplicity can be very helpful, just like  how Copernicus' simple circle model allowed Kepler and Newton to finally understand movement on Earth. More specifically, just like friction obstructed us from discovering the laws of mechanics here on Earth, more sophisticated features of language as well as all kinds of cultural aberrations may also obstruct us from seeing the foundational structures of meaning. There may of course also be more fundamental arguments for using pregroups, and Lambek took his conceptual motivation for pregroups from psychology.  Some computational linguists have strong feelings about which grammatical algebra to be used, and many think that Lambek got it wrong.  But even then,  Copernicus was wrong too, since planets don't move on circles around the sun, but without him we would not have been where we are now.   

%%%%%%%%%%%%%%%%%%%   IMPORTANT!!!   %%%%%%%%%%%%%%%%%%%
%
%A RELATED STORY IS WHY CLASSICAL PHYSICS IS NOT INTUITIVE.  THE GREEKS WHERE REALLY SMART AND DIDN'T COME UP WITH IT.  NEITHER DID ANYONE ELSE LATER BASED ON THE DATA OF OUR DAILY EXPERIENCES HERE ON EARTH.  WE HAD TO GO TO OUTER SPACE WHERE THERE IS NO FRICTION (VIA BRACHE'S DATA) IN ORDER TO COME UP WITH THE LAWS OF MECHANICS, SINCE HERE ON EARTH WE ARE DRIVEN TOWARDS NOTIONS LIKE ACTUAL/POTENTIAL.  LAGRANGIANS AND THE LIKE HAVE NO RELATION TO OUR DAILY EXPERIENCE, EXCEPT THEN MAYBE FOR POTENTIALS, WHICH BRINGS US BACK TO BASIC CONCEPT OF THE GREEKS. \e
%
%%%%%%%%%%%%%%%%%%%   IMPORTANT!!!   %%%%%%%%%%%%%%%%%%%

\par\medskip\noindent
{\bf Some related works.} The sentence type used in this paper was also used  in the recent DisCoCat paper that introduces Cartesian verbs \cite{CLM}, and as discussed in \cite{CLM} Sec.~2.3, precursors of this idea are in \cite{GrefSadr, Kartsaklis13reasoningabout, KartsaklisSadrzadeh2014}.

Also within the context of DisCoCat, the work by Toumi et al.~\cite{CDMT, AlexisMSc} involves multi-sentence interaction by relying on \em discourse representation structures \em \cite{kamp2013discourse}, which comes at the cost of reducing meaning to a number. %\TODOb{What exactly are the concessions/new ideas?}  
Textual context is also present in the DisCoCat-related papers \cite{polajnar2015exploration, wijnholds2018classical}, although no sentence composition mechanism is proposed.

Within more traditional natural language semantics research, \em dynamic semantics \em \cite{groenendijk1991dynamic, visser1998contexts}  models sentence meanings as I/O-transitions and text as compositions thereof. However, the approach is still rooted in predicate logic, just as Montague semantics is, hence not accounting for more general meaning spaces, and also doesn't admit the explicit type structure of diagrams/monoidal categories. Dynamic semantics is a precursor of \em dynamic epistemic logic (DEL) \em  \cite{BMS, CBS} which we briefly address in Sec.~\ref{sec:DEL}; we expect that DEL, and generalisations thereof,  may in fact emerge from our model of language meaning by considering an epistemics-oriented subset of meanings. In \cite{sadrzadeh2018static}, static and dynamic vector meanings are explicitly   distinguished, taking inspiration for the latter from dynamic semantics. There are many  other logic-oriented approaches to text e.g.~\cite{asher2003logics}, of text organisation e.g.~\cite{mann1988rhetorical}, and of the use of categorical structure.\footnote{Despite our somewhat provocative stance towards ML 
%in the paragraph entitled `The  need for structural understanding' in Sec.~\ref{sec:intro}, 
there is a clear scope for combining DisCoCirc with ML-methods.  Some work in this direction, for the case of DisCoCat, is \cite{MarthaNew}.}

\section{Background: DisCoCat}\label{sec:DisCoCat} 
 
\subsection{diagrams}\label{sec:diagrams} 
  
Diagrams are made up of  boxes: 
\ctikzfig{box}
each of which may have a number of inputs and outputs, and in this paper these boxes will typically be labeled by words or sentences.   Boxes without either inputs or outputs may also occur, which we call \em states \em and \em effects \em respectively, and we  depict them here as follows:
\ctikzfig{stateeffect}
The inputs and outputs of boxes can then be connected by wires yielding general \em diagrams \em e.g.: 
\ctikzfig{boxexample}
What determines a diagram are (see  also \cite{CKbook} Corollary 3.5 \& Definition 3.8):
\bit
\item the connectedness of the wire-structure, and
\item the labels on wires and boxes.
\eit

Diagrams can either be read  upward, like how we build structures from the ground up, or downward, like how gravity causes downfall.  I prefer the constructive view in which diagrammatic structures are built rather than where they emerge by letting the forces have a go.  Unfortunately, the elders of our portion of spoken language disagreed, and for that reason, in this paper, diagrams will be read from top to bottom. 

In category-theoretic terms, diagrams live in monoidal categories where the wires correspond to objects and the boxes correspond to morphisms.  Places where one can find easy-going introductions to the categories-diagrams connection include \cite{CatsII, CKbook}

There are two particular kinds of diagrams that will play a role in this paper, both discussed in great detail in \cite{CKbook}.
\par\medskip\noindent
{\bf Circuits} are diagrams obtained by composing boxes in parallel and sequentially.  In category-theoretic terms they live in a  \em symmetric monoidal category\em.  They admit a clear flow of time, from inputs to outputs, as a circuit carries a causal structure with boxes as nodes and wires as causal relationships (see   \cite{CKbook} Theorem 3.22). This in particular implies that an output of a box will always be connected to an input of a box in its `future'. 
%We will see that sentences in a text have the structure of such a circuit.
\par\medskip\noindent
{\bf String diagrams}  on the other hand  allow for outputs to be connected to any input, and even to other outputs.  Similarly,  inputs can be connected to inputs. This craziness is enabled by the fact that string diagrams allow for \em cap- and cup\em-shaped wires:
\[
\begin{tikzpicture}[scale=0.80]
	\begin{pgfonlayer}{nodelayer}
		\node [style=none] (0) at (1.25, -0.5) {};
		\node [style=none] (1) at (-1.25, -0.5) {};
	\end{pgfonlayer}
	\begin{pgfonlayer}{edgelayer}
		\draw [style=swap, bend left=90, looseness=1.50] (1.center) to (0.center);
	\end{pgfonlayer}
\end{tikzpicture}\qquad\qquad\qquad\begin{tikzpicture}[scale=0.80]
	\begin{pgfonlayer}{nodelayer}
		\node [style=none] (0) at (1.25, 0.5) {};
		\node [style=none] (1) at (-1.25, 0.5) {};
	\end{pgfonlayer}
	\begin{pgfonlayer}{edgelayer}
		\draw [style=swap, bend right=90, looseness=1.50] (1.center) to (0.center);
	\end{pgfonlayer}
\end{tikzpicture} 
\]
One can also think of these cap- and cup-shaped wires as boxes, respectively as a state and an effect, and then string diagrams can be given the shape of a circuit  (\cite{CKbook} Theorem 4.19).  Conveniently,  every one-input-one-output-box can be transformed into a two-output-state, and vice versa:  
\[
\input{MS1.tikz}\ \ \mapsto\ \ \input{MS2.tikz}\qquad\qquad\qquad\input{MS3.tikz}\ \ \mapsto\ \ \input{MS4.tikz}
\]
and this is a bijective correspondence called \em box-state duality \em (see \cite{CKbook} Sec.~4.1.2).
%This in particular implies that states on two systems are rich enough to capture all one-system boxes.  In quantum theoretical terms, this witnesses the presence of entanglement. 
%As already been shown in \cite{CSC} words in a sentence have the structure of a string diagram when we describe grammar by pregroups. 
More general uses of caps/cups for converting types are referred to as \em transposition\em.  In category-theoretic terms string diagrams live in a \em compact closed category \em (also called  \em autonomous category\em) \cite{KellyLaplaza, SelingerSurvey}.   

We also usually assume that string diagrams can be flipped vertically:
\[
\input{boxadj1.tikz}\ \ \mapsto \ \ \input{boxadj2.tikz}
\]
For example, if we flip a state, we get an effect, and vice versa, and if we flip a cap, we get a cup, and vice versa.  
 Above we also flipped the text in the boxes, but this was just to make a point, and won't do this anymore as this obstructs readability.  In category-theoretic terms this flipping is called a \em dagger \em structure, or \em adjoints \em \cite{AC2, SelingerCPM} (see also \cite{CKbook} Sec.~4.3.1).

 Another particular kind of box are \em spiders \em \cite{CQMII, CKbook}.  In category-theoretic terms they correspond to so-called dagger special commutative Frobenius algebras.  We represent them by a dot with some  input and output wires: 
\ctikzfig{spidercomp}
The key property of spiders is that they fuse together:  
\[
\input{spider.tikz}\ \ =\ \ \input{spidercomp.tikz}
\]
%This in particular means that any connected diagram made up of spiders only, is equal to any other such diagram if the number of input and output wires matches.  
An alternative way to think of these spiders is as \em multi-wires\em, which are generalised wires in that, rather than having two ends, can have multiple ends. What corresponds to fusion is that if two multi-wires are connected, then all the ends of one are also connected with all the ends of the other. 

As we will justify in Sec~\ref{sec:models}, the spider with three legs should be thought of as the logical AND:
\[
\input{AND1.tikz} \ \ :=\ \  \input{AND2.tikz} 
\]
The one with a single leg should be thought of as discarding.

\subsection{From grammar to wirings}\label{sec:grammarwire}

From the mathematics of sentence structure~\cite{Ajdukiewicz, Bar-Hillel, Lambek0} we know that the `fundamental particles' making up phrases and sentences are not  words, but some basic grammatical types. The noun-type $n$, and the type of whole sentences $s$ are examples of these basic types. On the other hand, the transitive-verb-type is not a basic type, but a composite  one made up of two noun-types and one sentence-type.  The precise manner in which these basic types interact depends on which \em categorial grammar \em one uses.  We will adopt Lambek's \em pregroups \em   \cite{Lambek1, LambekBook}, since it can be formulated  diagrammatically. 

For each basic type~$x$ there are two corresponding  `anti-types', which we denote~${}^{-1}x$ and~$x^{-1}$.  Think of these as a left and a right  inverse.  Then, in English, a transitive verb has type:
\[
{}^{-1}n \cdot s \cdot n^{-1}    
\]
To understand this type, consider a transitive verb like {\tt hate}. Simply saying {\tt hate} doesn't convey any useful information, until we also specify {\tt who  hates who}.  That's exactly the role of the anti-types: they specify that in order to form a meaningful sentence  the transitive verb needs a noun on the left and a noun on the right, which then cancel out with the anti-types:   
\[
\underbrace{\mbox{\tt Alice}}_{n} \underbrace{\mbox{\tt hates}}_{{}^{-1}n \cdot s \cdot n^{-1}}  \underbrace{\mbox{\tt Bob}}_n
\]  
So both $n \cdot {}^{-1}n$ and $n^{-1} \cdot n$ vanish, and what remains is $s$, confirming that 
{\tt Alice hates Bob} is a proper  sentence (a.k.a.~grammatically well-typed). 

So where are the diagrams? They depict the cancelations:     
\ctikzfig{hates}%\ctikzfig{hatestalk}
%
%
%\ctikzfig{s22romance}
%
%
%\ctikzfig{s22romance2}
%
%
For a more complex sentence (but with very similar meaning)  like: 
\begin{center}
{\tt Alice does not love Bob} 
\end{center}
the wiring will be more complex \cite{CSC}:
\ctikzfig{hatescomplgram}
but the idea remains the same.  In general, one can extract these wirings from the book \cite{LambekBook}, which assigns types to all grammatical roles. 

The main idea of DisCoCat is to think of these wires not just as a representation of an algebraic computation, but as a representation of how the meanings of the words making up the sentence interact.  Representing word-meanings as follows, also accounting for the types:
\ctikzfig{hates2}   
%where the boxes represent the meanings, and the outgoing wires correspond to the type of the word, 
we can now apply the wiring to these as follows: 
%\ctikzfig{hates3}   
\ctikzfig{s22}
The wires now `feed' the object and the subject into the verb in order to produce the meaning of the whole sentence.   
%As a result we obtain a representation of how meanings flow within a sentence, and this will become even more apparent in the following section.  

One should contrast this \em compositional \em model for word and sentence meanings to the \em bag-of-word-model \em still employed in distributional NLP and information retrieval \cite{harris1954distributional}, where as the name says, words in a sentence are treated just like a structureless bag. 

\subsection{Internal wirings of meanings}  \label{sec:wirings}

Not only grammatical structure, but also certain meanings  themselves can be represented using wiring. Examples include \em functional words\em, which play some kind of logical role and would cary no empirical data, and other words where a wiring structure provides a simplification. % in terms of the data-size. 
The internal structure of words can also help to better understand the overall meaning of a sentence. 

\subsubsection{Functional words} 

An example from \cite{PrelSadr, CSC} is the sentence:
\ctikzfig{hatescompl}
We see  that both {\tt does} and {\tt not} have been represented using wires: 
\ctikzfig{hatescomplclever}
{\tt does} being entirely made up of wires, while {\tt not} also involving a $\neg$-labeled box which represents negation of the meaning. %, something which obviously will not exist in every possible model of meaning, but does in some. 
That these wirings are well-chosen becomes clear when we yank them: 
\ctikzfig{s23}
i.e.~we get the negation of the meaning of  {\tt Alice loves Bob}.
%\ctikzfig{hatescompl2}
%where we also transposed {\tt like} in order to turn it into a function-like  box. 
%So we obtain the negation of {\tt like} applied to the pair ({\tt Alice}, {\tt Bob}), which is indeed the intended meaning of the sentence.
  
An example  from \cite{FrobMeanI, FrobMeanII} uses spiders for relative pronouns:
\ctikzfig{relpron}
Simplification now yields:  
\ctikzfig{s24} 
%\ctikzfig{relpron2}   
i.e.~the conjunction of {\tt she} (i.e.~being female) and the property {\tt [...]\!\!\! hates Bob}, which is indeed again the intended meaning of the sentence. 

%\TODOb{Should we pick up on this in the AND section?}
Building further on the idea that the merge spider represents AND, in \cite{DimitriCoord} an account was given of coordination between identical syntactic types.
 
\subsubsection{Adjectives and {\tt to be}}\label{sec:adj}
   
\em Intersective adjectives \em \cite{KampPartee1995} are adjectives which leave a noun unaltered except for specifying an additional property, e.g.~{\tt red car}, {\tt hot chilies} or {\tt sad Bob}, as opposed to {\tt crashed car},  {\tt rotten chilies} or {\tt dead Bob}. While a general adjective has a composite type e.g.:
\ctikzfig{s27}  
%\ctikzfig{adj1}  
the type of an intersective adjective can be reduced to a single wire \cite{ConcSpacI}:  
\ctikzfig{s28}  
%\ctikzfig{adj2}
yielding a conjunction: 
\ctikzfig{s25} 
%\ctikzfig{adj3}

Closely related to adjective is the verb {\tt to be}, since {\tt sad Bob} and {\tt Bob is sad} convey the same meaning. Of course, the overall  type of these two statements is different, being a noun and a sentence respectively, but we will see later that this difference vanishes when we move from DisCoCat to DisCoCirc.  Accepting {\tt Bob is sad} to be a noun, the following internal wiring of %`intersective uses' of 
the verb {\tt to be} is induced by the one of intersective adjectives: %\COMMb{Uses commutativity.}
\[
\input{s9.tikz}\ \ =\ \ \input{s10.tikz}\ \ =\ \ \input{s26.tikz}
\]

\subsubsection{Compact verbs}\label{sec:semicart}%\TODOb{Change name?  Even more restrictive than in paper.}
 
Recently, in \cite{CLM}, an internal wiring for a special type of verbs was proposed. The main idea is that the verb's only role is to impose an adjective {\tt verb}${}_s$ on the subject and an adjective {\tt verb}${}_o$ on the object. For example, {\tt paints} will put a paintbrush in the subject's hand, and makes the object change colour.  For many verbs this description can be used as an adequate first approximation.    So a transitive verb is  \em semi-Cartesian \em if it has the following internal wiring: 
\beq\label{eq:Cart}
\input{Cart.tikz}
\eeq
As indicated in the picture, this wiring  implies that its sentence type consists of two noun-types, which is a very natural choice to make. 

However,  a verb like {\tt being married to} is clearly not of that form, as it   expresses an entanglement of the specific subject and the specific object.  The natural generalisation of the idea of semi-Cartesian verbs is then:
\beq\label{eq:s34v2}
\input{s34v2.tikz}
\eeq
This representation was first used in \cite{GrefSadr}, and also in \cite{KartsaklisSadrzadeh2014}, and we call it the \em compact \em representation of transitive verbs. 

\subsection{Models of meaning}\label{sec:models}

Given that diagrams live in abstract (a.k.a.~axiomatic) categories, they allow for a wide range of concrete models.  It suffices to pick a  concrete category that has cups and caps, and then wires become the objects (a.k.a.~spaces) and boxes become the morphisms (a.k.a.~maps between these spaces).

In NLP, the \em vector space \em model takes  wires to be spaces of distributions and  boxes to be linear maps.  
%This model is also sometimes called , then thinking of these distributions as certain vectors in a vector space. 
The distributions are empirically established, by means of counting  co-occurences with a selected set of basis words 
%within a large body of text  
\cite{Schuetze}. % The basis vectors  then freely span the vector spaces.  
Adjusting this model to DisCoCat,  the cups and caps are:
\[
\begin{tikzpicture}[scale=0.80]
	\begin{pgfonlayer}{nodelayer}
		\node [style=none] (0) at (1.25, -0.5) {};
		\node [style=none] (1) at (-1.25, -0.5) {};
	\end{pgfonlayer}
	\begin{pgfonlayer}{edgelayer}
		\draw [style=swap, bend left=90, looseness=1.50] (1.center) to (0.center);
	\end{pgfonlayer}
\end{tikzpicture}\ \ :=\ \  \sum_i |i i\rangle\qquad\qquad\qquad\begin{tikzpicture}[scale=0.80]
	\begin{pgfonlayer}{nodelayer}
		\node [style=none] (0) at (1.25, 0.5) {};
		\node [style=none] (1) at (-1.25, 0.5) {};
	\end{pgfonlayer}
	\begin{pgfonlayer}{edgelayer}
		\draw [style=swap, bend right=90, looseness=1.50] (1.center) to (0.center);
	\end{pgfonlayer}
\end{tikzpicture}\ \ := \ \ \sum_i \langle i i|
\]
and spiders are in one-to-one correspondence with orthonormal bases \cite{CPV}. Explicitly,  given an orthonormal basis $\{ |i\rangle \}_i$ the spiders  arise as follows:
\[
\input{spidercomp.tikz}\ \ := \ \ \sum_i |i\ldots i\rangle\langle i\ldots i|
\]
Hence, caps and cups are instances of spiders, and so are \em copy \em and \em merge\em:  
\[
\input{copy.tikz}\ \ :=\ \  \sum_i |i i\rangle\langle i|\qquad\qquad\qquad\input{merge.tikz}\ \ := \ \ \sum_i |i\rangle\langle i i|
\]

Another model employed in DisCoCat instead considers \em sets and relations \em \cite{CSC}.  By thinking of relations as Boolean-valued matrices this model is closely related to the previous one, and can be thought of as `possibilistic' distributions (contra `probablilistic').   This particular model also justifies to interpret merge  as AND, as in the model states are subsets, and merge then corresponds to intersecting these states.

One can also take wires to be spaces of density matrices and boxes to be  superoperators. This model was initially introduced to account for ambiguity of meaning \cite{DimitriDPhil, RobinMSc, calco2015}, and was also used to capture lexical entailment \cite{EsmaSC, bankova2019graded}. It  will play a central role in this paper, although with a somewhat different interpretation. Again, density matrices can be established empirically \cite{calco2015}. 
%The spiders mentioned above cary over and have in the past been used for that purpose.  However, for a number of reasons, in this paper we will reconsider the use both of this specific model of spiders, as well as the spider structure itself (see Sec.~\ref{Sec:Model}).

Another recently developed model takes the convex subsets of certain convex spaces~\cite{ConcSpacI} to be the wires, following Gardenfors' conceptual spaces program \cite{gardenfors}. This model represents meanings in a manner that appeals directly to our senses.  A plethora of generalisations thereof are in \cite{DBLP:conf/wollic/CoeckeGLM17}.  Again, empirical methods can be used to establish meanings.

\subsection{Comparing  meanings}\label{sec:comparing} 

Once we have computed the meanings of sentences in a concrete model, we can compare these meanings.   Here are two examples of doing so: 
\par\medskip\noindent
{\bf Similarity.} Establishing similarity is one of the standard tasks in NLP, and one does this in terms of a distance-measure: the less the distance, the more similar meanings are.  
%In order to do that the spaces need to come with some notion of distance.  
One can use the inner-product (or some function thereof), which for sentences $\sigma_1$ or $\sigma_2$  is diagrammatically denoted as: 
\ctikzfig{inner}
as one can indeed think of an inner-product as the composition of one state with the adjoint of another state (cf.~Sec.~\ref{sec:diagrams}).  This manner of comparing meanings generalises to arbitrary dagger compact closed categories, such as the category of density matrices and completely positive maps.
%which in many cases can simply be taken to be the transpose (which exist in any compact closed category).  For example, using the presentation of density matrices as in \cite{CKbook} one can also use the inner-product.  
In the concrete representation of density matrices we obtain: 
\[
Tr\left( \sigma_2\sigma_1\right) \ = \ \ \input{innerHilb.tikz}\ \ = \ \ \input{innerHilb2.tikz}
\]
where we used the fact that density matrices are self-adjoint, and that the transpose of the adjoint is the conjugate (which is indicated by the bar).  
\par\medskip\noindent
{\bf Graded entailment.} One may want to know if one meaning entails another one.  Given the noisiness of empirical data, a useful strict entailment relation might be hard to achieve.  Instead, a graded entailment relation that tells us the degree to which one meaning entails another one is more useful.  Strict entailment relations correspond to partial orderings, and a graded ones correspond to a labeled extension thereof. Still, many models of meaning in use, like the vector space model,  don't even admit  a natural non-trivial graded entailment structure, and it is here that  density matrices have a role to play. As shown in \cite{bankova2019graded}, for those such a structure does exist and is well-studied, namely  the L\"owner ordering for positive matrices \cite{lowner}:
\[
\sigma_1 \leq_k \sigma_2\ \ \Leftrightarrow\ \ \sigma_2 - k \,\sigma_1\ \mbox{is positive}
\]
It is useful  to play around a bit  with the scaling of the density matrices. If one normalises density matrices by setting the trace to 1, then there are no strict comparisons.  On the other hand, when one sets the largest eigenvalue to 1, then we get for the specific case of  projectors (i.e.~scaled density matrices with all non-zero eigenvalues the same):  
\[
\sigma_1 \leq \sigma_2\ \ \Leftrightarrow\ \ \sigma_1 \circ \sigma_2 = \sigma_1
\]
just like in Birkhoff-von Neumann quantum logic \cite{BvN}, which is then naturally interpreted as  propositional inclusion.  Some alternative scalings are in \cite{van2017ordering}.

\section{Features and flaws of DisCoCat}\label{sec:featflaw}   

%Formally speaking, DisCoCat provides an algorithm that assigns a  meaning to sentences, given: 
%\ben
%\item the  meanings of its words, and, 
%\item its grammatical structure 
%\een
%Grammar is modelled by some categorial grammar, and word-meanings within some monoidal category that matches the structure of the grammar. % (cf.~compact closed categories match Lambek's pregroups). 
%\par\medskip\noindent 
Here is a summary of the   main features of DisCoCat:
\par\medskip\noindent 
{\bf Feature 1.} 
The initially  key identified   feature of DisCoCat was that meanings of sentences with different grammatical structure still live in the same space, something that  is crucial for comparing meanings (cf.~Sec.~\ref{sec:comparing}).  Earlier approaches that combined grammar and meaning, most notably, Smolensky's connectionist cognitive architecture \cite{SmolenskyBook}, did not have this feature.  The quest for a model that does so was put forward in \cite{ClarkPulman}.
%, which is the basis for tasks like search; note here that Google and its peers `don't understand grammar'. 
\par\medskip\noindent
{\bf Feature 2.} 
The DisCoCat algorithm that assigns meaning to sentences given the  meanings of its words and its grammatical structure 
 can be presented as an intuitive diagram that clearly shows how word-meanings  interact to produce the meaning of the sentence.
\par\medskip\noindent
{\bf Feature 3.} 
Wire-structure can be used in DisCoCat to provide meanings of functional words as we did in Sec.~\ref{sec:wirings} (while in standard NLP they are usually treated as noise), and to simplify the representation of words with composite types like adjectives and verbs as we did in  Secs.~\ref{sec:adj} and \ref{sec:semicart}.
\par\medskip\noindent
{\bf Feature 4.} While in this paper we used pregroups, as argued in \cite{LambekvsLambek, EdGDPhil}, DisCoCat also supports other categorial grammars such as standard Lambek calculus \cite{Lambek0}, Lambek-Grishin calculus \cite{grishin1983generalization} and CCG \cite{steedman2000syntactic}.
\par\medskip\noindent   
{\bf Feature 5.} 
As  discussed in Sec.~\ref{sec:models}, in DisCoCat word meanings can live in many kinds of spaces provided these organise themselves in a monoidal category that matches the structure of the grammar.
%In  follow-up work we moved away from the vector space models of NLP (which have a lot of `uglyness' and `wrongs') about them to richer models such as density matrices that allow for encoding: {\bf a.}~word ambiguity \cite{DimitriDPhil, RobinMSc, calco2015} just like mixedness does that  in quantum theory, and {\bf b.}~entailment via the L\"owner order \cite{bankova2016graded}.  We also crafted a model based G\"ardenfors's conceptual spaces which more closely resemble human senses-based  thought \cite{ConcSpacI}.  
\par\medskip\noindent 
{\bf Feature 6.} 
DisCoCat allows  for integrating grammar and meaning in one whole. In the above we indeed had examples of compositional structure entering meaning-boxes, which then interact with the grammatical structure.\footnote{Admittedly, more work needs to be done for further exploiting this feature. Crucially,  while initial formulations of DisCoCat either used a categorical product or a functor in order to combine grammar and meaning \cite{CSC, PrelSadr}, the way forward is to assume a compositional structure encompassing both grammar and aspects of meaning.} 
%This is necessary as there are many instances where meaning informs sentence structure and vice versa. Initial formulations of DisCoCat either used a categorical product or a functor in order to combine grammar and meaning \cite{CSC, PrelSadr}.  Instead, one can work with a concrete category that has the relevant structure, and then there is flexibility in what to consider to be grammar and what to be meaning.\e
\par\medskip\noindent
{\bf Feature 7.} 
Contra the bag-of-words model in NLP \cite{harris1954distributional}, the spaces for different grammatical types  vary, which reflects the fact that their functionality within sentences is very different.  In other words, meaning spaces are typed, with all the usual advantages.  If words can play different grammatical roles, e.g.~both as noun and adjective, then there also are canonical ways for interconverting these, e.g.~a noun becomes an adjective as follows:
\[
\input{red1.tikz}\ \ \mapsto\ \ \input{red2.tikz} 
\]
\par\medskip\noindent
{\bf Feature 8.} 
Proof-of-concept experiments showed that DisCoCat outperformed its competitors for certain academic benchmark tasks \cite{GrefSadr, KartSadr}.
\par\medskip\noindent
As already indicated in the introduction, DisCoCat has some  shortcomings: 
\par\medskip\noindent
{\bf Flaw 1.} DisCoCat does not answer the question of how the meanings of sentences compose in order to provide the meaning of an entire text.
%, which obstructs  many practical applications where  the meaning of entire texts or conversations is relevant. 
\par\medskip\noindent
{\bf Flaw 2.} DisCoCat   assumes words to have a fixed meaning, while in text meanings will typically evolve.
%many practical applications are about learning and information gain, and we may also want to use meanings as  representation of our knowledge, or others' knowledge, and learn as more data comes in.  
\par\medskip\noindent
{\bf Flaw 3.} DisCoCat doesn't determine the sentence type. 
\par\medskip\noindent
%{\bf Flaw 4.} A practical issue with DisCoCat was the size of the vectors needed for words like transitive verbs with a composite type, which quickly becomes huge.  Therefore,  some hacks are needed,  but a conceptually founded approach 
We will now resolve each of these flaws  in one go!

\section{Composing sentences: meet DisCoCirc}

We will represent the $|\sigma_i|$ words in a sentence $\sigma_i$ as a horizontal string, and the $|\tau|$ sentences in a text $\tau$ as a vertical stack: 
\[\begin{array}{lcl}
{\tt Word}^1_1 & \ldots & {\tt word}^{|\sigma_1|}_1.\\  
&{\tt.}&\\
&{\tt.}&\\
&{\tt.}&\\
{\tt Word}^1_{|\tau|} & \ldots & {\tt word}^{|\sigma_n|}_{|\tau|}.
\end{array}\]
 
\subsection{Naive composition of sentences for DisCoCat} 

In DisCoCat, each of the sentences (cf.~those in Sec.~\ref{sec:DisCoCat}) is a state, i.e.~they have a single output of sentence-type, and no input.  This substantially restricts the manner in which we can compose them.   
%Here we wish to compose sentences, just like we do when forming a narrative, and see how words-meanings interact in order to form the meaning of the entire narrative. 
The structure available to us in DisCoCat %, as outlined in the previous section, 
are wires and spiders. %, which can be used to represent conjunction.  
The desirable thing to do is to  also rely on this  structure for composing sentences, so that word-meaning composition can  interact with sentence-meaning composition.  
%If these would be two non-interacting compositional dimensions then we obviously achieved nothing.  
But then, pretty much the only thing one can do is to take the conjunction of all sentences:   
\beq\label{diag:sentencebag}
\input{s1.tikz} 
\eeq
%where $\sigma_1, \sigma_2 \ldots \sigma_n$ are sentences, we used vertical ordering to represent  the order of the sentences in a piece of text.  
%In analogy with the bag-of-words model \cite{harris1954distributional}, 
We could call this the \em bag-of-sentences model\em,  %\TODOb{Not true if we move to a non-commutative conjunction, so if we keep we should stipulate that we are using the commutative conjunction of DisCoCat, which could then later be overturned since DisCoCircuit needs non-commutative updating.}
%Diagram (\ref{diag:sentencebag}) represents indeed  a `bag of sentences', 
since without changing the diagram we can flip the  vertical order of  sentences,  
%mix the  vertical order up in any other way
or not even give one:
\[
\input{s2.tikz}
\qquad\qquad\qquad\qquad
\input{s3.tikz}
\]
%(here the horizontal ordering is of course also entirely interchangeable, but this is insignificant as we agreed that sentence ordering is reflected by vertical ordering)  
An example where this makes perfect sense is:
\begin{center}
%{\tt The sun shines.\\ 
{\tt It is cloudy.\\ 
%The grass is green.\\
Liverpool has beaten Napoli.\\
%The rabbits are hopping around.}
Brexit has become a total mess.} 
\end{center}
as these sentences clearly commute.  
%This is in particular the case given they have nothing to do with each other whatsoever.  Even when the sentences are related e.g.~joint describing a scenery:
%\TODOb{Here the example involving a big mood change is hidden.}
%However, although  sentences may  commute in terms of facts established when we consider them as one whole, we do loose something when swapping sentences, which becomes more apparent in the following example: 
%\begin{center}
%{\tt The sun shines.\\
%The grass is green.\\
%The rabbits are hopping.\\
%A plane drops a bomb. 
%}
%\end{center}
A  non-example is:
\begin{center}
{\tt Bob is born.\\
Bob drinks beer.\\
Bob dies. 
}
\end{center}
Here one could still  argue that the meaning of the sentences now dictates their ordering, so the latter could be extracted even if the sentences arrive in a bag, of course, at the cost of having to know the  meanings of all words.  The argument completely breaks down here:
\begin{center}
{\tt 
Add egg yolk and salt.\\
Whisk mix for 20 seconds.\\
Add mustard and acid.\\
Whisk mix for 30 seconds.\\
Slowly add oil while whisking.\\
}
\end{center}
as the order of adding ingredients is key to making good mayonnaise.  Changing the order still would result in a meaningful recipe, but not mayonnaise. 

More importantly, what is also clear from this example is that a lot more is going on  besides the order of things: the ingredients and actions interact with each other similarly to how words in a sentence interact with each other as described in Sec.~\ref{sec:DisCoCat}.  Our  bag-of-sentence-model doesn't reflect any of that.  

We will now make that interaction structure explicit, while still only making use of the structure available to us in DisCoCat.  Of course, something will need to change, and 
%this will start with revisiting the representation of sentences.  so 
rather than a conservative extension of the DisCoCat framework, we introduce a fundamentally modified framework, DisCoCirc, while retaining the  features  of the DisCoCat framework  listed in Sec.~\ref{sec:featflaw}.

%This ingredients, of course are again going to be wires and the like representing flows of meanings.

%What we loose here is the dramatic build-up.  
%\bR In order to retain it in the structural representation we need to add something that prevents commuting, while still retaining the ability to evaluate that the  overall factual outcome is the same. \bR We will show how to do this further below\e, as an instance of an in general non-commutative model of sentence composition that we introduce now, and which will involve a modification of the representation of sentences themselves.  
%
%DRAMA CAN BE ADDED AS A `mood'-SYSTEM THAT GETS AFFECTED BY THE SENTENCES, AND HENCE PREVENTS COMMUTATION. \e

\subsection{Sentences as I/O-processes}\label{sec:main}

Consider the following example:  
\begin{center}
{\tt Alice 
is a dog.\\
Bob is a person.\\
Alice bites Bob.} 
\end{center}
Clearly, the meaning of the third sentence crucially depends on what we learn about the meaning of the nouns {\tt Alice} and {\tt Bob} in the first two sentences, turning {\tt dog bites man} into {\tt man bites dog} if {\tt Bob} were to be a {\tt dog} and {\tt Alice} were to be a {\tt person}.
Also, before the 1st sentence is stated, {\tt Alice} is just a meaningless name, and the same goes for {\tt Bob} until the 2nd sentence is stated.  
So the meaning of  {\tt Alice} and {\tt Bob} evolves as the text progresses,  
and it is the sentences that update our knowledge about {\tt Alice} and {\tt Bob}.

What we propose is that the 3rd sentence, which  would look like:
\ctikzfig{s4}
 in DisCoCat, would instead be drawn like this:
\ctikzfig{s5}
%\bR From now on, the states representing nouns will become  wires labeled  by the \em character \em they represent, and  we will take the sentence type to also comprise these two noun-wires (and possibly others): \e
So in particular, the nouns {\tt Alice} and {\tt Bob} are now not states but wires (a.k.a~types) and the sentence is an I/O-box: 
\beq\label{eq:sentasbox}
\input{s6.tikz}
\eeq
with the nouns {\tt Alice} and {\tt Bob} both as inputs and as outputs. In this way, the sentence can act on the nouns and update their meanings.
%Note that the latter matches the sentence-type of the semi-Cartesian verbs  discussed in Sec.~\ref{sec:wirings}.  
Hence:  
\begin{center}
\fbox{\em A sentence is not a state, but a process,\em}  
\end{center} 
that represents how  words in it are updated by that sentence. 

For the remainder of this paper we will restrict ourselves to updating nouns, but the same applies to other word-types. 
Using the wire-representation of the verb {\tt to be} of Sec.~\ref{sec:adj}, the wire-representation of {\tt Alice is a dog} becomes:
\[
\input{s17.tikz}\ \ =\ \ \input{s11.tikz}
\]
and similarly, that of {\tt Bob is a person} becomes: 
\ctikzfig{s12} 
%We could still take some nouns to be static attributes, for example {\tt person} and {\tt dog}, and then the first two  sentences simply can be depicted as, respectively:
In Sec.~\ref{sec:wirings} we mentioned that while our treatment of {\tt to be} leads to a noun-type, this wouldn't be a problem  anymore in DisCoCirc. And indeed, within our new sentences-as-processes realm we obtain the same  type as the  sentence (\ref{eq:sentasbox}) simply by adjoining a `passive' wire: 
\[
\input{s13.tikz}\qquad\qquad\qquad\mbox{}\qquad\qquad\qquad\input{s14.tikz} 
\]
This passive wire stands for the fact that the ancillary noun is part of the text as a whole, but doesn't figure in this particular sentence.  By adding it, the I/O-types of all the sentences are the same, so they can be composed:
%
%Altogether the composition of the three sentences now becomes: 
\beq\label{eq:s15NON-CART}
\input{s15NON-CART.tikz}
\eeq
%\ctikzfig{s15}%CARTESIAN VERSION
So in general, given a text, we end up with a wire diagram that looks like this: 
%for sentences in I/O-shape as explained above:
\beq\label{sentencecircuit}
\input{s19.tikz}
\eeq
where the sentences themselves also have a wire diagram.  In particular, it's a process, and this process alters our understanding of words in the text.
This yields another slogan:
\begin{center}
\fbox{\em Text is a process that alters meanings of words\em} 
\end{center} 

Using the form (\ref{eq:s34v2}) for {\tt biting} we obtain for {\tt Alice bites Bob}:
\beq\label{commabuse2}
\input{s7v2.tikz}\ \ = \ \ \input{s8v2.tikz}
\eeq
%It may be the case that one only cares that the subject bites, meaning `being a dangerous dog', and/or that the object has been bitten, meaning `having to go to the hospital', then we could use the semi-Cartesian verb form (\ref{eq:Cart}):
%\ctikzfig{s8}
%In order to further simplify diagram (\ref{eq:s15NON-CART}) and for the sake of clarity,  we will take {\tt bites} to be a semi-Cartesian verb, then yielding the following diagram for the 3rd sentence: %\COMMb{Uses commutativity (or we could put a twist for {\tt B}${}_o$).}
%%,  and (for now) ignoring whatever new feature it introduces: 
%\beq%\label{commabuse2}
%\tikzfig{s7}\ \ = \ \ \tikzfig{s8}
%\eeq
%%(here {P}${}_s$ and {P}${}_o$ respectively stand for something along the lines of {\tt putting teeth into something} and {\tt experiencing penetrating teeth}) 
Diagram  (\ref{eq:s15NON-CART}) now simplifies to:   
\beq\label{DogBitesMeaning} 
\input{s16v2.tikz}
\eeq
and is clearly distinct from the case of {\tt man bites dog},  which would be: 
\ctikzfig{s18v2}
%In the case of the semi-Cartesian verb form we obtain:
%\[
%\tikzfig{s16}\ \ \not= \ \ \tikzfig{s18}
%\]
%where the inequality expresses that `being a dangerous dog' and `having to go to the hospital' are very different attributes.

One thing we can also now do now is  show that different texts can have the same meaning.  For example, 
the single sentence: 
\begin{center}
{\tt 
Alice who is a dog bites Bob who is a person.   
}
\end{center} 
% results in  wire-diagram (\ref{DogBitesMeaning}),  so diagrams can expose that different texts have the same meaning.  Spelling out both sentence and text wire-structure, it should also be clear that the diagram:
which has as its diagram:
\ctikzfig{s21v2}
can, using spider-fusion, be directly `morphed' into the diagram:
\ctikzfig{s20v2} 

One thing  we may be aiming for is a network that shows how the different nouns 
are related, or maybe just whether  they either are related, or not at all.  
How the network is connected will depend on, for example, the kinds of verbs that appear in the text, and which subject-object pairs they connect.  A further simplification can be made if we only require a binary knowledge, e.g.~{\tt knows} vs.~{\tt doesn't know},  which we can respectively represent as:
\[
\input{knows.tikz}\qquad\qquad\qquad\qquad\input{doesntknow.tikz}
\]  
In the resulting network we then get  clusters of connected nouns.  

\subsection{The use of states}\label{sec:states}  

In the main example of Sec.~\ref{sec:main} we made use of states to represent {\tt dog} and {\tt person}.  The reason we  use states for them is that the text doesn't help us understand these nouns. In contrast, the text is all about helping us  understand {\tt Alice} and {\tt Bob} and their interactions.  So we can clearly distinguish two roles for nouns:
\bit 
\item \em Static nouns\em: the text \underline{does not} alter our understanding of them.
\item \em Dynamic nouns\em: the text \underline{does}  alter our understanding of them.
\eit
This distinction between dynamic and static nouns may seem somewhat artificial, and indeed, it exists mainly for practical purposes.  From a foundational perspective the natural default would be to let all nouns be dynamic, and not just nouns but all words, since also adjectives and verbs may be subject to change of meaning.  However, taking only some nouns to be static is a very reasonable simplification given that  in a typical text the meaning of many other words would not alter in any significant manner.  Doing so significantly simplifies diagrams, and in particular their width.  This does give rise to the practical question of how to decide on the `cut' between the  dynamic and static nouns. We briefly  address this in Sec.~\ref{sec:meaning}.  %For certain task the set of dynamic nouns may actually be a given,  as we also discuss in Sec.~\ref{sec:meaning}.\TODOb{This should be expended and maybe moved to a dedicated tasks section, summarising stuff like in the network section.}    
 
Also in the main example of Sec.~\ref{sec:main}, we had no prior understanding of {\tt Alice} nor {\tt Bob}.\footnote{This is in particular the case given that the  names {\tt Alice} and {\tt Bob} are gender neutral (cf.~Alice Cooper and Bob in Blackadder II \& IV).}  In general we may already have some prior understanding about certain nouns.  One way to specify this is by means of  \em initial states\em, to which we then apply the circuit representing the text:%\TODOb{Do we really want a dot here?}
\[
\input{s39.tikz}
\]
where the initial states represented by plain dots stand for the case of no prior understanding (cf.~{\tt Alice} and {\tt Bob} earlier). Without changing the circuit we can also put them where they enter the text:  
\[
\input{s40.tikz}
\]
%In these circuits, , so among dynamic nouns we can still distinguish two cases: 
%\bit
%\item \ \,\tikzfig{s41}\ \,: we have \underline{no prior understanding}
%\item \tikzfig{s42}: we have \underline{some prior understanding}
%\eit

Of course, once we insert initial states, we cannot precompose with other text anymore.  A straightforward way to avoid this problem is by instead using \em initial processes\em:
\ctikzfig{s45}
which adjoin understanding to a type just like adjectives do.

\subsection{DisCoCat from DisCoCirc}\label{Sec:CatFromCirc}

We now show that DisCoCat is an instance of DisCoCirc.  Assuming that (1) text is restricted to a single sentence,
and that (2) nouns are static, we exactly obtain DisCoCat sentences.  Alternatively, assuming that (2') dynamic nouns have an initial state, we  also obtain a  DisCoCat sentence: 
\[
\input{s43.tikz} \ \ = \ \ \input{s44.tikz} 
\]
%Hence, DisCoCat is a special case of DisCoCirc.

\subsection{Individual and subgroup meanings}\label{sec:subgroup}  

In many cases one would just be interested in the meaning of a single dynamic noun, rather than the global meaning of a text.  Or, maybe one is interested in the specific relation of two or more dynamic nouns. %\TODOb{We need a shorter name for these e.g.~character.}  
The way to achieve this is by discarding all others.  For example, here we  care about the meaning of the 2nd dynamic noun:
\ctikzfig{s56} 
while here we care about the relationship between the 2nd and the 5th one: 
\ctikzfig{s57bis}
The latter can for example teach us if two agents either agree or disagree, or, either cooperate or anti-cooperate. In the case of three or more agents it can tell us more refined forms of interaction, e.g.~do they pairwise cooperate or globally (cf.~the W-state vs.~the GHZ-state in quantum theory \cite{DurVC}).
%\TODOb{This should be picked up again in concrete calc.~as this avoids calculating huge overall meaning.}   

Subgroups may also arise naturally, when agents vanish from the story e.g.~by being murdered.  In that case  discarding instead of being of an epistemic nature is actual  ontic vanishing:
\ctikzfig{s57}
This vanishing can even be a part of the verb structure for those verbs that induce the vanishing of an object, for example:
\beq\label{eq:kill}
\input{kills1v2.tikz} %\ \ = \ \ \tikzfig{kills2}
\eeq
Of course, if it remains of importance who the actual killer is, then we shouldn't use the simple semi-Cartesian verb structure.

%\bR ... adjectives may refine specification of an agents and this indicates that there is a subset structure on agents that should be considered. This can arise by network generation (see movie example), making it implicit in the diagram ... \e

\subsection{Example}\label{sec:example} 

Consider the following text:\footnote{Loosely adapted from ``C'era una volta il West'', Sergio Leone, 1968.}
\begin{center}
{\tt Harmonica (is the brother of) Claudio.\\  
Frank hangs Claudio.\\
Snaky (is in the gang of) Frank.\\
Harmonica shoots Snaky.\\ 
Harmonica shoots Frank.} 
\end{center}
As a diagram this becomes: 
\ctikzfig{West1}
Using spider-fusion, transposition of states into effects, and identifying 2-legged spiders with either caps, cups or plain wires, simplifies this to: 
\beq\label{cowboydiagram}
\input{West2.tikz} 
\eeq
Notice that  one party induces the effect, while the other one is subjected to termination, matches the grammatical subject-object distinction.
There are indeed  two dimensions to diagram (\ref{cowboydiagram}), a static one,  representing the connections between the dynamic nouns:
\ctikzfig{West5}
as well as  a temporal-causal structure associated to these. 

\subsection{Other cognitive modes}

The mathematical formalism presented here for text structure may be equally useful for modelling other cognitive modes, not just the linguistic one.  One obvious example is the visual  mode, which we can think of as movies.  Here dynamic nouns correspond to the characters of the movie, and sentences to scenes.  The grammatical structure then corresponds to interactions of characters. For example, the scene:
  \begin{center}
\epsfig{figure=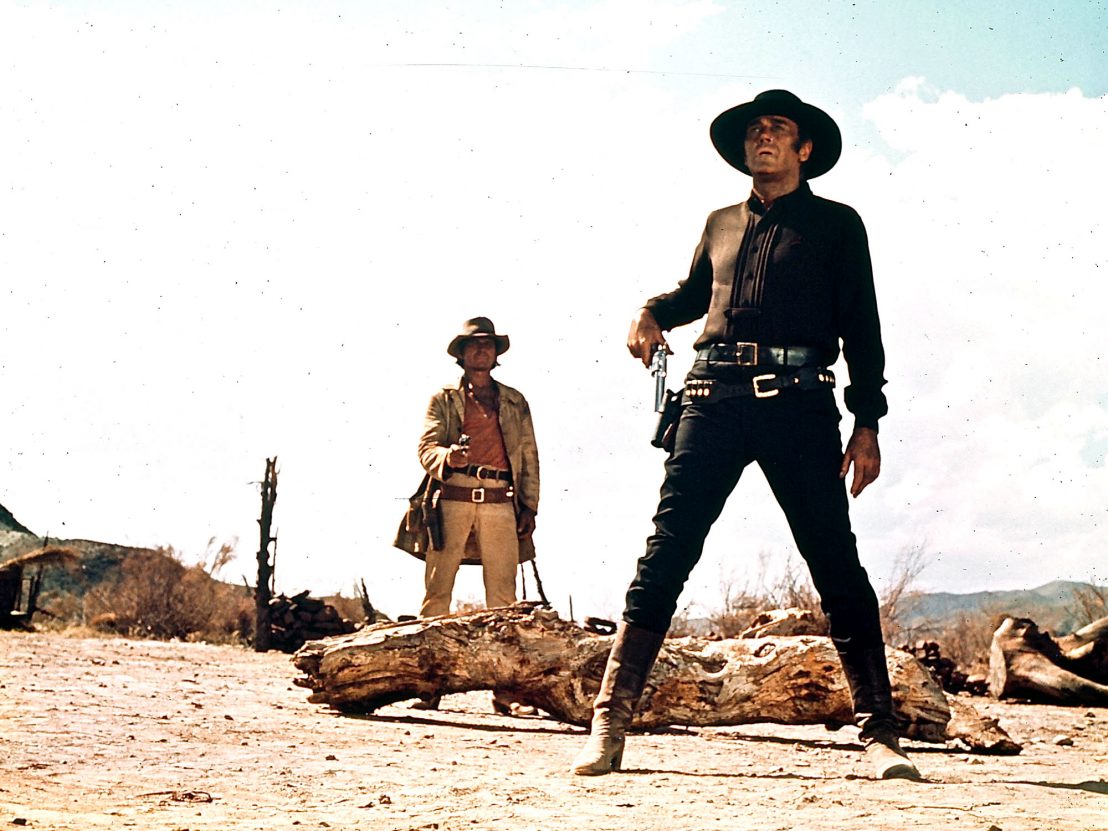,width=180pt}
  \end{center}
corresponds to the sentence:
\begin{center}
{\tt Harmonica shoots Frank.} 
\end{center}
The subject corresponds to  Harmonica (played by Bronson), the object to Frank (played by Fonda) and the verb is the shooting of Frank by Harmonica:   
\ctikzfig{West4}
More broadly the verb corresponds to the interaction of the characters.  Text corresponds to sequences of scenes, which `act' on the characters that take part in it, hence forming a circuit. 

Having a matching diagrammatic formalism for text and for movies  allows one then to make  translations between these, via the corresponding diagrams.
For example, we can translate the example of Sec.~\ref{sec:example} to a movie:    
\[
%\hspace{-2mm}\tikzfig{West1'}
\raisebox{0mm}{\mbox{\LARGE(\ref{cowboydiagram})}}\ \ \ \quad \mbox{\LARGE$\mapsto$}\quad\ \  \input{West3.tikz} 
\]
where the snapshots represent the entire scene they are part of.

\section{Logic and language}

Propositional logic emerged from language, translating words like {\tt and}, {\tt or} and {\tt not} into logical connectives AND, OR and NOT. under the impetus of Aristotle and others.  DisCoCirc re-enforces that link with several branches of modern non-classical logics. Here are two proof-of-concept examples.

\subsection{Dynamic epistemic logic from language}\label{sec:DEL} 
%\TODOb{This further splits up the role of dynamics into ontic and epistemic change.}

Epistemic logic is  concerned with how one represents knowledge in logical terms, and \em dynamic epistemic logic \em (DEL) \cite{BMS, CBS} how this logic gets updated when acquiring new knowledge, e.g.~from communication, using language.  Since in DisCoCirc we have a build-in update mechanism, one may expect that DEL-update could emerge from DisCoCirc-update, and hence DEL could directly emerge from language structure.   This seems indeed to be the case.  
 
In order to establish this,  the types {\tt Alice} and {\tt Bob} will now represent the knowledge of those \em agents\em, rather than what we know about them. Sentences describing communication of knowledge typically involves a doubly transitive verb (i.e.~one that both has a direct and an indirect object),   
or alternatively, a preposition like {\tt to}. For example:
\begin{center}
{\tt Alice tells Bob (a) secret.}
\end{center}
\begin{center}
{\tt Alice tells (a) secret to Bob.}  
\end{center}
As we haven't proposed a wiring yet for a doubly transitive verb nor for {\tt to}, we will do so now, and we will also give an internal wiring for {\tt tells}  specific to this epistemic context. Grammatical wirings are taken from \cite{LambekBook}.  Setting:   
\[
\input{DEL1.tikz}
\]
\[
\input{DEL2.tikz}
\]
indeed result in the same simplified diagram: %\TODOb{Clear non-commutativity; resolving requires twists in either {\tt to} or {\tt tells}.}
\beq\label{commabuse} 
\input{DEL4.tikz}\ \ = \ \ \input{DEL3.tikz} 
\eeq
From this then follows an obvious wiring of {\tt knows}: 
%(or {\tt believes}) within this epistemic context: 
\ctikzfig{DEL5}
This is  the same wiring as we had for {\tt is}, which makes sense, since {\tt being} in an ontic context translates to {\tt knowing} in an epistemic context. 
 
We hope to  further develop this link in a dedicated forthcoming paper, being guided by the conviction 
that a diagrammatic framework for DEL can be established that directly draws from spoken language, and that moreover allows ust to accommodate a wide variety of models beyond the propositional and probabilistic ones.   

\subsection{Linear and non-linear {\tt and}}\label{Sec:AND}  

%\TODOb{Something should be mentioned here in relation to Lambek too, somewhere.} 
Another feature of  DisCoCirc  is that it dictates   different  representations of \underline{\tt and}, namely, when either conjoining properties that subjects possess, or,  conjoining the subjects themselves.  
%The same is the case when either having \underline{\tt and}   within an direct object or within a indirect object.  
In linear logic (LL) lingo \cite{Girard, Seely}, these  respectively correspond to a \em linear conjunction \em and a \em non-linear conjunction\em. Hence, these different representations correspond to  uses of \underline{\tt and} with  different meanings.  Consider the sentence:
\begin{center}
{\tt Alice wears (a) hat \underline{and} (a) scarf.} 
\end{center}
and the sentence:
\begin{center}
{\tt Alice \underline{and} Bob wear (a) hat.}
\end{center}
The fundamental difference between these sentences is that:
\bit
\item Alice wearing both a hat and a scarf only requires \underline{one} Alice, while,
\item Alice and Bob both wearing a hat requires \underline{two} hats. 
\eit
In the case of the former we assign two properties to a single agent, namely {\tt wearing a hat} and {\tt wearing a scarf}, while in the case of the latter a single property,  {\tt wearing a hat},  is attributed to two agents.  This means that we require (a.k.a~`consume' in LL lingo) the property {\tt wearing a hat} twice, hence, it needs to be copied. 
%and in particular, we will be needing two hats. On the other hand, for Alice to wear both a hat and a scarf we obviously don't need two Alices.  
Put differently, non-linear conjunction allows for copying, so it is the AND we have in classical propositional logic.  

A physics analogy would be that \underline{and} in the 1st sentence refers to having two physical particles e.g.~a {\tt proton \underline{and}  electron}, while  \underline{and} in the 2nd sentence lists two properties of a single particle, e.g.~{\tt position \underline{and}  velocity}.   

The non-linear \underline{and} is what we have been using until now all the time in this paper by means of spiders.  So it is the linear \underline{and} that needs an alternative treatment. We will do what is  standard  when representing two things in a string diagrams (see \cite{CKbook} Section 3.1.1),  namely  putting  two wires side-by-side.  So the different representations of AND look as follows:
\[
\input{s50.tikz}\qquad\qquad  \input{s49.tikz}
\]
Then we get for the 1st sentence:
\[
\input{s51.tikz}
\]
and for the 2nd sentence we instead have: 
\[
\input{s52.tikz}
\]
where internally in {\tt wear} some copying must happen:
\[
\input{s53.tikz}
\]
The difference is also apparent in how each of the sentences can be decomposed in two sentences, which in the case of the 1st one yields a sequential composition, and in the case of the 2nd one a parallel composition:
\[
\input{s54.tikz}\qquad\qquad\input{s55.tikz}
\]
%In a sentence useing both conjections, we can use bialgebra to swap copy and AND e.g. Alice and Bob wear hat and scarf.

\section{Concrete models}

We gave the beginnings of a compositional structure of word and sentence composition, with (multi-)wires and boxes as primitives, and illustrated how one reasons with these in the absence of concrete models.  We now provide some ideas for which kinds of models are particularly suitable for DisCoCirc.

\subsection{Sketch of a concrete model}\label{Sec:Model} 

As we are dealing with updating and corresponding information gains, the  vector space model of NLP (see Section \ref{sec:models}) won't do.
Density matrices 
do have a clear notion of information gain, 
and for this reason  form the basis of quantum information theory (see e.g.~\cite{BennettShor}).  We now describe the ingredients of a DisCoCirc model based on density matrices.
States, e.g.~word meanings:
\ctikzfig{state}
are  density matrices.
An example of a state is the maximally mixed state of quantum theory, which has a density matrix corresponding to the (scaled) identity, and represents the state of no information whatsoever.  
\ctikzfig{dell}
The states of perfect information correspond to the pure states of quantum theory, that is, matrices arising as doubled vectors $|\psi\rangle\langle\psi|$.  The effect:
\ctikzfig{del}
corresponds to the trace.  General processes, e.g.~sentence meanings: 
\ctikzfig{box}
correspond to trace-preserving completely positive maps. Pure processes are those that send pure states to pure states, and these arise from linear maps as the Krauss forms $f^\dagger\circ - \circ f$. In this manner, the spiders: 
\ctikzfig{spidercomp}
of Sec.~\ref{sec:models} become part of this model too.  

However, there are reasons to move away from these particular choices of spiders, and even part of the axioms for spiders. We give some suggestions here of some potential alternatives.  The cups and caps of Sec.~\ref{sec:models} remain perfectly ok, so they respectively become:
\[
\begin{tikzpicture}[scale=0.80]
	\begin{pgfonlayer}{nodelayer}
		\node [style=none] (0) at (1.25, -0.5) {};
		\node [style=none] (1) at (-1.25, -0.5) {};
	\end{pgfonlayer}
	\begin{pgfonlayer}{edgelayer}
		\draw [style=swap, bend left=90, looseness=1.50] (1.center) to (0.center);
	\end{pgfonlayer}
\end{tikzpicture}\ \ :=\ \  \sum_{ij} |i i\rangle\langle jj |\qquad\qquad\quad\begin{tikzpicture}[scale=0.80]
	\begin{pgfonlayer}{nodelayer}
		\node [style=none] (0) at (1.25, 0.5) {};
		\node [style=none] (1) at (-1.25, 0.5) {};
	\end{pgfonlayer}
	\begin{pgfonlayer}{edgelayer}
		\draw [style=swap, bend right=90, looseness=1.50] (1.center) to (0.center);
	\end{pgfonlayer}
\end{tikzpicture}\ \ := \ \ \sum_{ij} \langle i i|\circ-\circ| jj \rangle
\]

The main role of merge is to assign properties. One could set:  
\beq\label{eq:projadj}
\input{Proj1.tikz}\ \ :=\ \ P_{\tt x}\circ - \circ P_{\tt x} 
\eeq
where $P_{\tt x}$ is the orthogonal projector on the subspace corresponding to property {\tt x}. As suggested in \cite{widdows2003word, Widdows} it is indeed  natural to think of subspaces as properties, just like in Birkhoff-von Neumann quantum logic \cite{BvN} which we already mentioned earlier. This operation inherits  associativity from diagram composition, which is an obvious minimal requirement.  It is not commutative, but that also makes perfect sense when thinking of the changing colours of a chameleon, where post-composition should discard previous colours. After re-scaling projectors become a special case of density matrices, and using spectral decomposition $\rho_{\tt x}=\sum_i p_i P_i$ one can associate properties to general density matrices, for example as follows:
\[
\input{Proj1.tikz} \ \ :=\ \ \sum_i p_i P_i\circ - \circ P_i   
\] 
In the follow-up paper \cite{CoeckeMeich} we present a class of similar generalisations.

%For certain purposes one may however desire a commutative multiplication, e.g.~for modelling AND as in Sec.~\ref{Sec:AND}. One could either use the usual spiders, or symmetrise the non-commutative multiplication.   Staying with logical connectives, it has been suggested that orthogonal projection of subspaces could represent NOT \cite{Widdows}, and this naturally extends to density matrices as making the difference with the scaled identity. 

\subsection{Computing text meaning}\label{sec:meaning}

%\subsection{Algorithmic realisation of the text meaning} %\TODOb{Put refs to relevant sections and illustrating pics.}

The following steps  produce the data needed to derive text meaning:
\ben
\item Identify sentences using punctuation.
\item Establish  grammatical types of all words  using standard parsers.   
\item Identify the dynamic nouns.  As this concerns a  new concept, this is also a new task and hence will need additional research. One could rely to some extent on  grammar and multiplicity of occurrence throughout the text, but   actual meaning will likely also play a role.   
%\bR Possibly, machine learning may help here.  \e
For certain problems the dynamic nouns may be a given, namely those that are of particular interest as part of the statement of the problem, for example, when analysing the relationship of certain parties of particular interest.

\item Form a diagram (see Sec.~\ref{sec:main}):
\bit
\item The dynamic nouns are the systems of the circuit.
\item The sentences are the gates of the circuit.
\item The internal wiring of the gates is given by the grammar.
\eit  
\item Establish meanings of states, which can be done using standard methods, or those previously developed for DisCoCat.
%and establish meanings for other grammatical types.  Some work has been done in the direction of the latter, but more is needed, in particular when using representations of verbs with internal wirings.
\een
In order to obtain the actual meaning of the text:
\ben\setcounter{enumi}{5}
\item Insert all meanings into the diagram.
\item For computing the resulting (possibly simplified) diagram, one way to do so is to decompose the diagram in tensor products and sequential compositions of boxes, caps/cups and dot operations.  A more direct manner for computing diagrams is outlined in Theorem 5.61 of \cite{CKbook}.  
\een
%\subsection{Computing text meanings}

\subsection{Comparing texts}  

To compare texts we can simply rely on what we did in DisCoCat (see Sec.~\ref{sec:comparing}), provided we use initial states (see Sec.~\ref{sec:states}) so that the meaning of the text as a whole becomes a state, or,  compute similarity as follows: 
\ctikzfig{innerproc}  
using a generalisation of the Hilbert-Schmidt product, i.e.~the inner-product applied to the states arising from box-state duality (see Sec.~\ref{sec:diagrams}):  
\begin{equation}\label{eq:map-state}
\input{map-state.tikz} 
\end{equation}
Also graded entailment is obtained as in DisCoCat (see Sec.~\ref{sec:comparing}) when using initial states, or, representation  text as a state as in  (\ref{eq:map-state}).
%\bR ...we have L\"owner but this depends on the availability of a minus-sign for density matrices, which may not exist for either propositional representations or visual ones. In that case we can define an entailment relation using spiders ... HOW EXACTLY ...Task: compare to L\"owner... \e 

\section{Physical embodiment}

In the abstract we mentioned that while the developments in this paper are independent of a physical embodiment, most notably a classical vs.~a quantum embodiment, both the compositional formalism and the suggested concrete  model of meaning of Section \ref{Sec:Model} are highly \em quantum-inspired\em. The compositional structure is directly imported from quantum theory \cite{teleling, Gospel, DBLP:books/daglib/p/Coecke17}, the suggested concrete  model of meaning employs the density matrices which von Neumann designed specifically for quantum theory \cite{vNdensity}, and also our suggested alternatives for spiders belong to an area of current activity in quantum foundations (see e.g.~\cite{Leifer1, leifer2013towards, CoeckeSpekkens2012, PhysRevX.7.031021, JonSina}), which aims for a quantum analog of Bayesian inference theory. 
 
Therefore, it should come as no surprise that implementation of DisCoCirc on a quantum computer would come with a wide range of benefits.  For example, as pointed out in \cite{WillC}, classically the required space resources grow exponentially in the number of dynamic nouns, and this exponential growth could vanish on a quantum computer. Similarly, density matrices substantially increase the space required to represent meanings, while for a quantum computer they come for free.     Regarding time resources, quantum computational speed-ups have already been identified for DisCoCat  \cite{WillC}, by exploiting progress in quantum machine learning \cite{wiebe2012quantum}, and these straightforwardly carry over to DisCoCirc.  Expect many dedicated publications on further advantages of implementing DisCoCirc on a quantum computer to be forthcoming, and in fact, \em quantum natural language processing \em (QNLP) may become one of the leading areas of the so-called NISQ era \cite{preskill2018quantum}, given its tolerance for imperfection \cite{WillC}.  Currently, efforts are under way to implement the quantum algorithm of \cite{WillC} on a simulator \cite{Intel}, and very recently, another paper appeared \cite{Wiebe-Smolensky} that is entirely dedicated to QNLP. 

\section*{Thanks}  

We thanks   Dan Marsden, Dusko Pavlovic and Alexis Toumi for valuable discussions on the content of the paper, including DM contributing ``The gloves are off!" slogan as his interpretation of the DisCo-ing cats cartoon. Phil Scott also provide corrections and suggestions.  The  SYCO 3, QPL and ACT referees also provided useful feed-back including pointers to related work, which was also provided by Valeria de Paiva, Graham White, Patrik Eklund and Alexandre Rademaker.  We are grateful to Ilyas Khan for the additional  motivational context within which this paper was produced.

\bibliographystyle{plain}
\bibliography{main}

\begin{thebibliography}{10}

\bibitem{AC1}
S.~Abramsky and B.~Coecke.
\newblock A categorical semantics of quantum protocols.
\newblock In {\em Proceedings of the 19th Annual IEEE Symposium on Logic in
  Computer Science (LICS)}, pages 415--425, 2004.
\newblock {a}rXiv:quant-ph/0402130.

\bibitem{AC2}
S.~Abramsky and B.~Coecke.
\newblock {Abstract physical traces}.
\newblock {\em Theory and Applications of Categories}, 14(6):111--124, 2005.
\newblock arXiv:0910.3144.

\bibitem{Ajdukiewicz}
K.~Ajdukiewicz.
\newblock Die syntaktische konnexit\"at.
\newblock {\em Studia Philosophica}, 1:1--27, 1937.

\bibitem{PhysRevX.7.031021}
J.-M.~A. Allen, J.~Barrett, D.~C. Horsman, C.~M. Lee, and R.~W. Spekkens.
\newblock Quantum common causes and quantum causal models.
\newblock {\em Phys. Rev. X}, 7:031021, 2017.

\bibitem{asher2003logics}
.~Asher and A.~Lascarides.
\newblock {\em Logics of conversation}.
\newblock Cambridge University Press, 2003.

\bibitem{EsmaSC}
E.~Balkir, M.~Sadrzadeh, and B.~Coecke.
\newblock {\em Distributional Sentence Entailment Using Density Matrices},
  pages 1--22.
\newblock Springer International Publishing, Cham, 2016.

\bibitem{CBS}
A.~Baltag, B.~Coecke, and M.~Sadrzadeh.
\newblock Algebra and sequent calculus for epistemic actions.
\newblock {\em Electronic Notes in Theoretical Computer Science}, 126:27--52,
  2005.

\bibitem{BMS}
A.~Baltag, L.S. Moss, and S.~Solecki.
\newblock The logic of public announcements, common knowledge, and private
  suspicions.
\newblock In {\em Proceedings of the 7th conference on Theoretical aspects of
  rationality and knowledge}, pages 43--56. Morgan Kaufmann Publishers Inc.,
  1998.

\bibitem{bankova2019graded}
Dea Bankova, Bob Coecke, Martha Lewis, and Dan Marsden.
\newblock Graded hyponymy for compositional distributional semantics.
\newblock {\em Journal of Language Modelling}, 6(2):225--260, 2019.

\bibitem{Bar-Hillel}
Y.~Bar-Hillel.
\newblock A quasiarithmetical notation for syntactic description.
\newblock {\em Language}, 29:47--58, 1953.

\bibitem{BennettShor}
C.~H. Bennett and P.~W. Shor.
\newblock Quantum information theory.
\newblock {\em IEEE Trans. Inf. Theor.}, 44(6):2724--2742, September 2006.

\bibitem{BvN}
G.~Birkhoff and J.~von Neumann.
\newblock The logic of quantum mechanics.
\newblock {\em Annals of Mathematics}, 37:823--843, 1936.

\bibitem{ConcSpacI}
J.~Bolt, B.~Coecke, F.~Genovese, M.~Lewis, D.~Marsden, and R.~Piedeleu.
\newblock Interacting conceptual spaces {I}.
\newblock In M.~Kaipainen, A.~Hautam\"aki, P.~G\"ardenfors, and F.~Zenker,
  editors, {\em Concepts and their Applications}, Synthese Library, Studies in
  Epistemology, Logic, Methodology, and Philosophy of Science. Springer, 2018.
\newblock to appear.

\bibitem{teleling}
S.~Clark, B.~Coecke, E.~Grefenstette, S.~Pulman, and M.~Sadrzadeh.
\newblock A quantum teleportation inspired algorithm produces sentence meaning
  from word meaning and grammatical structure.
\newblock {\em Malaysian Journal of Mathematical Sciences}, 8:15--25, 2014.
\newblock arXiv:1305.0556.

\bibitem{ClarkPulman}
S.~Clark and S.~Pulman.
\newblock Combining symbolic and distributional models of meaning.
\newblock In {\em Proceedings of AAAI Spring Symposium on {Q}uantum
  {I}nteraction}. AAAI Press, 2007.

\bibitem{Kindergarten}
B.~Coecke.
\newblock Kindergarten quantum mechanics.
\newblock In A.~Khrennikov, editor, {\em Quantum Theory: Reconsiderations of
  the Foundations III}, pages 81--98. AIP Press, 2005.
\newblock {a}rXiv:quant-ph/0510032.

\bibitem{Gospel}
B.~Coecke.
\newblock An alternative {G}ospel of structure: order, composition, processes.
\newblock In C.~Heunen, M.~Sadrzadeh, and E.~Grefenstette, editors, {\em
  Quantum Physics and Linguistics. A Compositional, Diagrammatic Discourse},
  pages 1 -- 22. Oxford University Press, 2013.
\newblock arXiv:1307.4038.

\bibitem{DBLP:books/daglib/p/Coecke17}
B.~Coecke.
\newblock From quantum foundations via natural language meaning to a theory of
  everything.
\newblock In S.~B. Cooper and M.~I. Soskova, editors, {\em The Incomputable:
  Journeys Beyond the Turing Barrier}, Theory and Applications of
  Computability, pages 63--80. Springer International Publishing, 2017.
\newblock arXiv:1602.07618.

\bibitem{CDMT}
B.~Coecke, G.~De~Felice, D.~Marsden, and A.~Toumi.
\newblock Towards compositional distributional discourse analysis.
\newblock In M.~Lewis, B.~Coecke, J.~Hedges, D.~Kartsaklis, and D.~Marsden,
  editors, {\em {\rm Proceedings of the 2018 Workshop on} Compositional
  Approaches in Physics, NLP, and Social Sciences, {\rm Nice, France, 2-3rd
  September 2018}}, volume 283 of {\em Electronic Proceedings in Theoretical
  Computer Science}, pages 1--12. Open Publishing Association, 2018.

\bibitem{DBLP:conf/wollic/CoeckeGLM17}
B.~Coecke, F.~Genovese, M.~Lewis, and D.~Marsden.
\newblock Generalized relations in linguistics and cognition.
\newblock In Juliette Kennedy and Ruy J. G.~B. de~Queiroz, editors, {\em Logic,
  Language, Information, and Computation - 24th International Workshop, WoLLIC
  2017, London, UK, July 18-21, 2017, Proceedings}, volume 10388 of {\em
  Lecture Notes in Computer Science}, pages 256--270. Springer, 2017.

\bibitem{LambekvsLambek}
B.~Coecke, E.~Grefenstette, and M.~Sadrzadeh.
\newblock Lambek vs. {L}ambek: Functorial vector space semantics and string
  diagrams for {L}ambek calculus.
\newblock {\em Annals of Pure and Applied Logic}, 164:1079--1100, 2013.
\newblock arXiv:1302.0393.

\bibitem{CQMII}
B.~Coecke and A.~Kissinger.
\newblock Categorical quantum mechanics {II}: Classical-quantum interaction.
\newblock {\em International Journal of Quantum Information}, 14(04):1640020,
  2016.

\bibitem{CKbook}
B.~Coecke and A.~Kissinger.
\newblock {\em Picturing Quantum Processes. A First Course in Quantum Theory
  and Diagrammatic Reasoning}.
\newblock Cambridge University Press, 2017.

\bibitem{CLM}
B.~Coecke, M.~Lewis, and D.~Marsden.
\newblock Internal wiring of cartesian verbs and prepositions.
\newblock In M.~Lewis, B.~Coecke, J.~Hedges, D.~Kartsaklis, and D.~Marsden,
  editors, {\em {\rm Proceedings of the 2018 Workshop on} Compositional
  Approaches in Physics, NLP, and Social Sciences, {\rm Nice, France, 2-3rd
  September 2018}}, volume 283 of {\em Electronic Proceedings in Theoretical
  Computer Science}, pages 75--88. Open Publishing Association, 2018.

\bibitem{CoeckeMeich}
B.~Coecke and K.~Meichanetzidis.
\newblock Meaning updating of density matrices.
\newblock {\em arXiv:2001.00862}, 2020.

\bibitem{CatsII}
B.~Coecke and {\'E}.~O. Paquette.
\newblock Categories for the practicing physicist.
\newblock In B.~Coecke, editor, {\em New Structures for Physics}, Lecture Notes
  in Physics, pages 167--271. Springer, 2011.
\newblock {a}rXiv:0905.3010.

\bibitem{CPV}
B.~Coecke, D.~Pavlovi{\'c}, and J.~Vicary.
\newblock A new description of orthogonal bases.
\newblock {\em Mathematical Structures in Computer Science, to appear},
  23:555--567, 2013.
\newblock {a}rXiv:quant-ph/0810.1037.

\bibitem{CSC}
B.~Coecke, M.~Sadrzadeh, and S.~Clark.
\newblock Mathematical foundations for a compositional distributional model of
  meaning.
\newblock In J.~van Benthem, M.~Moortgat, and W.~Buszkowski, editors, {\em A
  Festschrift for Jim Lambek}, volume~36 of {\em Linguistic Analysis}, pages
  345--384. 2010.
\newblock ar{x}iv:1003.4394.

\bibitem{CoeckeSpekkens2012}
B.~Coecke and R.~W. Spekkens.
\newblock Picturing classical and quantum bayesian inference.
\newblock {\em Synthese}, 186(3):651--696, 2012.

\bibitem{cybenko1989approximations}
G.~Cybenko.
\newblock Approximations by superpositions of a sigmoidal function.
\newblock {\em Mathematics of Control, Signals and Systems}, 2:183--192, 1989.

\bibitem{DurVC}
W.~D{\"{u}}r, G.~Vidal, and J.~I. Cirac.
\newblock Three qubits can be entangled in two inequivalent ways.
\newblock {\em Physical Review A}, 62(062314), 2000.

\bibitem{gardenfors}
P.~G{\"a}rdenfors.
\newblock {\em The Geometry of Meaning: Semantics Based on Conceptual Spaces}.
\newblock MIT Press, 2014.

\bibitem{Girard}
J.~Y. Girard.
\newblock {Linear logic}.
\newblock {\em Theoretical Computer Science}, 50(1):1--101, 1987.

\bibitem{EdGDPhil}
E.~Grefenstette.
\newblock {\em Category-Theoretic Quantitative Compositional Distributional
  Models of Natural Language Semantics}.
\newblock PhD thesis, University of Oxford, 2013.

\bibitem{GrefSadr}
E.~Grefenstette and M.~Sadrzadeh.
\newblock Experimental support for a categorical compositional distributional
  model of meaning.
\newblock In {\em The 2014 Conference on Empirical Methods on Natural Language
  Processing.}, pages 1394--1404, 2011.
\newblock ar{X}iv:1106.4058.

\bibitem{grishin1983generalization}
V.~N. Grishin.
\newblock On a generalization of the {A}jdukiewicz-{L}ambek system.
\newblock {\em Studies in nonclassical logics and formal systems}, pages
  315--334, 1983.

\bibitem{groenendijk1991dynamic}
J.~Groenendijk and M.~Stokhof.
\newblock Dynamic predicate logic.
\newblock {\em Linguistics and philosophy}, 14(1):39--100, 1991.

\bibitem{harris1954distributional}
Z.~S. Harris.
\newblock Distributional structure.
\newblock {\em Word}, 10(2-3):146--162, 1954.

\bibitem{KampPartee1995}
H.~Kamp and B.~Partee.
\newblock Prototype theory and compositionality.
\newblock {\em Cognition}, 57:129--191, 1995.

\bibitem{kamp2013discourse}
H.~Kamp and U.~Reyle.
\newblock {\em From discourse to logic: Introduction to modeltheoretic
  semantics of natural language, formal logic and discourse representation
  theory}, volume~42.
\newblock Springer Science \& Business Media, 2013.

\bibitem{DimitriDPhil}
D.~Kartsaklis.
\newblock {\em Compositional Distributional Semantics with Compact Closed
  Categories and Frobenius Algebras}.
\newblock PhD thesis, University of Oxford, 2014.

\bibitem{DimitriCoord}
D.~Kartsaklis.
\newblock Coordination in categorical compositional distributional semantics.
\newblock In {\em Semantic Spaces at the Intersection of NLP, Physics and
  Cognitive Science}, 2016.
\newblock arXiv:1606.01515.

\bibitem{KartSadr}
D.~Kartsaklis and M.~Sadrzadeh.
\newblock Prior disambiguation of word tensors for constructing sentence
  vectors.
\newblock In {\em The 2013 Conference on Empirical Methods on Natural Language
  Processing.}, pages 1590--1601. ACL, 2013.

\bibitem{KartsaklisSadrzadeh2014}
D.~Kartsaklis and M.~Sadrzadeh.
\newblock A study of entanglement in a categorical framework of natural
  language.
\newblock In {\em Proceedings of the 11th Workshop on Quantum Physics and Logic
  (QPL)}. Kyoto ‚Japan, 2014.

\bibitem{Kartsaklis13reasoningabout}
D.~Kartsaklis, M.~Sadrzadeh, S.~Pulman, and B.~Coecke.
\newblock Reasoning about meaning in natural language with compact closed
  categories and {F}robenius algebras.
\newblock In {\em Logic and Algebraic Structures in Quantum Computing and
  Information}. Cambridge University Press, 2015.
\newblock arXiv:1401.5980.

\bibitem{KellyLaplaza}
G.~M. Kelly and M.~L. Laplaza.
\newblock Coherence for compact closed categories.
\newblock {\em Journal of Pure and Applied Algebra}, 19:193--213, 1980.

\bibitem{Lambek0}
J.~Lambek.
\newblock The mathematics of sentence structure.
\newblock {\em American Mathematics Monthly}, 65, 1958.

\bibitem{lambek1968deductive}
J.~Lambek.
\newblock Deductive systems and categories.
\newblock {\em Mathematical Systems Theory}, 2(4):287--318, 1968.

\bibitem{lambek1969deductive}
J.~Lambek.
\newblock Deductive systems and categories ii. standard constructions and
  closed categories.
\newblock In {\em Category theory, homology theory and their applications I},
  pages 76--122. Springer, 1969.

\bibitem{lambek1980lambda}
J.~Lambek.
\newblock From lambda-calculus to cartesian closed categories.
\newblock {\em To HB Curry: essays on combinatory logic, lambda calculus and
  formalism}, pages 375--402, 1980.

\bibitem{lambek1993categorial}
J.~Lambek.
\newblock From categorial grammar to bilinear logic.
\newblock {\em Substructural logics}, 2:207--237, 1993.

\bibitem{Lambek1}
J.~Lambek.
\newblock Type grammar revisited.
\newblock {\em Logical Aspects of Computational Linguistics}, 1582, 1999.

\bibitem{LambekBook}
J.~Lambek.
\newblock From word to sentence.
\newblock {\em Polimetrica, Milan}, 2008.

\bibitem{LambekLNP}
J.~Lambek.
\newblock Compact monoidal categories from linguistics to physics.
\newblock In B.~Coecke, editor, {\em New Structures for Physics}, Lecture Notes
  in Physics, pages 451--469. Springer, 2011.

\bibitem{Leifer1}
M.~S. Leifer and D.~Poulin.
\newblock Quantum graphical models and belief propagation.
\newblock {\em Annals of Physics}, 323(8):1899--1946, 2008.

\bibitem{leifer2013towards}
M.~S. Leifer and R.~W. Spekkens.
\newblock {Towards a formulation of quantum theory as a causally neutral theory
  of {Bayesian} inference}.
\newblock {\em Physical Review A}, 88(5):052130, 2013.

\bibitem{MarthaNew}
M.~Lewis.
\newblock Compositionality for recursive neural networks.
\newblock arXiv:1901.10723, 2019.

\bibitem{lowner}
K.~L{\"o}wner.
\newblock {\"U}ber monotone matrixfunktionen.
\newblock {\em Mathematische Zeitschrift}, 38(1):177--216, 1934.

\bibitem{mann1988rhetorical}
W.~C. Mann and S.~A. Thompson.
\newblock Rhetorical structure theory: Toward a functional theory of text
  organization.
\newblock {\em Text-Interdisciplinary Journal for the Study of Discourse},
  8(3):243--281, 1988.

\bibitem{Intel}
Intel Newsroom.
\newblock Intel to support the {I}rish centre for high end computing on new
  collaborative quantum computing project, 2019.
\newblock
  https://newsroom.intel.ie/news-releases/intel-to-support-the-irish-centre-for-high-end-computing-on-new-collaborative-quantum-computing-project/.

\bibitem{RobinMSc}
R.~Piedeleu.
\newblock Ambiguity in categorical models of meaning.
\newblock Master's thesis, University of Oxford, 2014.

\bibitem{calco2015}
R.~Piedeleu, D.~Kartsaklis, B.~Coecke, and M.~Sadrzadeh.
\newblock Open system categorical quantum semantics in natural language
  processing.
\newblock In {\em CALCO 2015}, 2015.
\newblock ar{X}iv:1502.00831.

\bibitem{polajnar2015exploration}
T.~Polajnar, L.~Rimell, and S.~Clark.
\newblock An exploration of discourse-based sentence spaces for compositional
  distributional semantics.
\newblock In {\em Proceedings of the First Workshop on Linking Computational
  Models of Lexical, Sentential and Discourse-level Semantics}, pages 1--11,
  2015.

\bibitem{PrelSadr}
A.~Preller and M.~Sadrzadeh.
\newblock Bell states and negative sentences in the distributed model of
  meaning.
\newblock {\em Electronic Notes in Theoretical Computer Science},
  270(2):141--153, 2011.

\bibitem{preskill2018quantum}
J.~Preskill.
\newblock Quantum computing in the nisq era and beyond.
\newblock {\em Quantum}, 2:79, 2018.

\bibitem{Redei1}
M.~Redei.
\newblock Why {J}ohn von {N}eumann did not like the {H}ilbert space formalism
  of quantum mechanics (and what he liked instead).
\newblock {\em Studies in History and Philosophy of Modern Physics},
  27(4):493--510, 1996.

\bibitem{FrobMeanI}
M.~Sadrzadeh, S.~Clark, and B.~Coecke.
\newblock The {F}robenius anatomy of word meanings {I}: subject and object
  relative pronouns.
\newblock {\em Journal of Logic and Computation}, 23:1293--1317, 2013.
\newblock ar{X}iv:1404.5278.

\bibitem{FrobMeanII}
M.~Sadrzadeh, S.~Clark, and B.~Coecke.
\newblock The {F}robenius anatomy of word meanings {II}: possessive relative
  pronouns.
\newblock {\em Journal of Logic and Computation}, 26:785--815, 2016.
\newblock arXiv:1406.4690.

\bibitem{sadrzadeh2018static}
M.~Sadrzadeh and R.~Muskens.
\newblock Static and dynamic vector semantics for lambda calculus models of
  natural language.
\newblock {\em arXiv:1810.11351}, 2018.

\bibitem{JonSina}
S.~Salek and J.~Barrett.
\newblock Quantum weak causal modelling, 2019.
\newblock Draft.

\bibitem{Schuetze}
H.~Sch{\"u}tze.
\newblock Automatic word sense discrimination.
\newblock {\em Computational linguistics}, 24(1):97--123, 1998.

\bibitem{Seely}
R.~A.~G. Seely.
\newblock Linear logic, {$*$}-autonomous categories and cofree algebras.
\newblock {\em Contemporary Mathematics}, 92:371--382, 1989.

\bibitem{SelingerCPM}
P.~Selinger.
\newblock Dagger compact closed categories and completely positive maps.
\newblock {\em Electronic Notes in Theoretical Computer Science}, 170:139--163,
  2007.

\bibitem{SelingerSurvey}
P.~Selinger.
\newblock A survey of graphical languages for monoidal categories.
\newblock In B.~Coecke, editor, {\em New Structures for Physics}, Lecture Notes
  in Physics, pages 275--337. Springer-Verlag, 2011.
\newblock {a}rXiv:0908.3347.

\bibitem{SmolenskyBook}
P.~Smolensky and G.~Legendre.
\newblock {\em The Harmonic Mind: From Neural Computation to
  Optimality-Theoretic Grammar Vol.~{I}: Cognitive Architecture Vol.~{II}:
  Linguistic and Philosophical Implications}.
\newblock MIT Press, 2005.

\bibitem{steedman2000syntactic}
Mark Steedman.
\newblock {\em The syntactic process}, volume~24.
\newblock MIT press Cambridge, MA, 2000.

\bibitem{AlexisMSc}
A.~Toumi.
\newblock Categorical compositional distributional questions, answers \&
  discourse analysis.
\newblock Master's thesis, University of Oxford, 2018.

\bibitem{van2017ordering}
J.~van~de Wetering.
\newblock Ordering information on distributions.
\newblock {\em arXiv:1701.06924}, 2017.

\bibitem{visser1998contexts}
A.~Visser.
\newblock Contexts in dynamic predicate logic.
\newblock {\em Journal of Logic, Language and Information}, 7(1):21--52, 1998.

\bibitem{vNdensity}
J.~von Neumann.
\newblock Wahrscheinlichkeitstheoretischer aufbau der quantenmechanik.
\newblock {\em Nachrichten von der Gesellschaft der Wissenschaften zu
  G{\"o}ttingen, Mathematisch-Physikalische Klasse}, 1:245--272, 1927.

\bibitem{vN}
J.~von Neumann.
\newblock {\em Mathematische grundlagen der quantenmechanik}.
\newblock Springer-Verlag, 1932.
\newblock Translation, {\it Mathematical foundations of quantum mechanics},
  Princeton University Press, 1955.

\bibitem{Widdows}
D.~Widdows.
\newblock Orthogonal negation in vector spaces for modelling word-meanings and
  document retrieval.
\newblock In {\em 41st Annual Meeting of the Association for Computational
  Linguistics}, Japan, 2003.

\bibitem{widdows2003word}
D.~Widdows and S.~Peters.
\newblock Word vectors and quantum logic: Experiments with negation and
  disjunction.
\newblock {\em Mathematics of language}, 8(141-154), 2003.

\bibitem{Wiebe-Smolensky}
N.~Wiebe, A.~Bocharov, P.~Smolensky, M.~Troyer, and K.~M. Svore.
\newblock Quantum language processing.
\newblock {\em arXiv:1902.05162}, 2019.

\bibitem{wiebe2012quantum}
N.~Wiebe, D.~Braun, and S.~Lloyd.
\newblock Quantum algorithm for data fitting.
\newblock {\em Physical review letters}, 109(5):050505, 2012.

\bibitem{wijnholds2018classical}
G.~Wijnholds and M.~Sadrzadeh.
\newblock Classical copying versus quantum entanglement in natural language:
  The case of vp-ellipsis.
\newblock {\em arXiv:1811.03276}, 2018.

\bibitem{WillC}
W.~Zeng and B.~Coecke.
\newblock Quantum algorithms for compositional natural language processing.
\newblock arXiv:1608.01406.

\end{thebibliography}

\end{document}